\documentclass{article}
\usepackage[utf8]{inputenc}
\usepackage[main,final]{neurips_2026}
\usepackage{graphicx}
\usepackage{colortbl}
\usepackage{xcolor}
\definecolor{gaincolor}{RGB}{0,150,80}
\definecolor{dropcolor}{RGB}{190,70,50}

\usepackage[most]{tcolorbox}
\tcbuselibrary{listings,breakable,skins}
\tcbuselibrary{listings}
\newtcblisting{promptbox}[1]{
    title = {#1},
    breakable,
    enhanced,
    colback = white,
    colframe = black!15,
    coltitle = black,
    fonttitle = \bfseries,
    sharp corners,
    boxrule = 0.6pt,
    left = 5pt,
    right = 5pt,
    top = 4pt,
    bottom = 4pt,
    listing only,
    listing options = {
        basicstyle = \ttfamily\footnotesize,
        breaklines = true,
        columns = fullflexible,
        keepspaces = true,
        showstringspaces = false,
        tabsize = 2
    }
}

\usepackage[utf8]{inputenc} % allow utf-8 input
\usepackage[T1]{fontenc}    % use 8-bit T1 fonts
\usepackage{float}          % LAYOUT REPAIR: load float before hyperref
\usepackage{hyperref}       % hyperlinks
\usepackage{url}            % simple URL typesetting
\usepackage{booktabs}       % professional-quality tables
\usepackage{longtable}      % appendix mechanism tables with repeated headers
\usepackage{tabularx}       % tables constrained to the text width
\usepackage{placeins}       % keep appendix tables with their discussion
\usepackage{amsfonts}       % blackboard math symbols
\usepackage{nicefrac}       % compact symbols for 1/2, etc.
\usepackage{microtype}      % microtypography
\usepackage{xcolor}         % colors

\usepackage{algorithm}
\usepackage{algorithmic}
\usepackage{amsmath}
\usepackage{threeparttable}
\usepackage{amsthm}

\usepackage{amssymb}

\author{Zijie Xing\thanks{Equal contribution.} \ , Xingyue Liu\footnotemark[1] \ , Runze Wang, Luoming Hu \& Yanming Shen\thanks{Corresponding author.} \\
Dalian University of Technology\\
{\small \texttt{\{kokojjj, aroma,  runze\_wang, huluoming2004\}@mail.dlut.edu.cn, shen@dlut.edu.cn}} \\
}

\begin{document}

\makeatletter
\renewcommand{\@noticestring}{}
\makeatother

\title{KnowEvo: Knowledge Evolution for Protein Optimization}

\maketitle
\begin{abstract}
Large language models (LLMs) have shown promise in scientific discovery, yet systematically accumulating and refining reusable scientific knowledge remains challenging. Existing approaches primarily improve LLM-based problem solving through parameter optimization or feedback-driven refinement of prompts, programs, and solution trajectories, leaving acquired expertise either embedded in model weights or scattered across iterative problem-solving processes. As a result, knowledge is difficult to systematically inspect, evaluate, and refine. We propose \emph{KnowEvo}, a framework that treats externalized knowledge itself as the optimization target. 
Our key insight is that scientific knowledge is inherently modular: multiple interacting mechanisms jointly determine the design outcomes. This makes knowledge evolution fundamentally challenging, as improvements cannot be attributed to individual components in
isolation. To address this, \emph{KnowEvo} represents knowledge as executable expertise
blocks and organizes their refinements into strategy mutation trees, enabling
structured exploration over a combinatorial knowledge space.
The framework combines a Thompson-style forward allocation strategy that adaptively distributes evaluation budgets across strategy variants with a backward revision process, in which a code agent operates in an isolated sandbox to translate rollout evidence into localized program updates. This design enables persistent accumulation, evaluation, and refinement of mechanistic expertise across iterative experimentation.
We evaluate \emph{KnowEvo} on protein optimization tasks including stability and solubility. Results show that knowledge evolution consistently improves over initial strategy blocks and outperforms monolithic program evolution and prior LLM-guided evolution baselines across metrics such as pass@$k$, MRR, MAP, and NDCG@5. Code is available at \url{https://github.com/recynie/KnowEvo}.

\end{abstract}
\section{Introduction}
% \YM{We need a name for our method, how about this ``KnowEvo: Knowledge Evolution for Protein Optimization" }
% \YM{
% \begin{itemize}
%     \item key idea: optimizing externalized, persistent, reusable knowledge instead of model parameters.
%    \item Why it is challenging? (i) How to transform unstructured textual knowledge into a structured and optimizable space? Block-based mutation trees turn knowledge into structured, evolvable search space. (ii)  Context rot: How to manage vast trajectories from executions? Agentic workspace. (iii) How to convert sparse, global evaluation signals into structured, fine-grained updates for knowledge refinement? Decomposed reward\ZJ{the GFP and solubility tasks dont have decomposed reward, maybe it is hard to claim decomposed reward is one of the major innovation } and Thompson Sampling with Depth
% Bonus.
% \item Difference to RL/GRPO
% \item Why not RL? (i) Updating model weights is insanely expensive; (ii) when a model is trained to be highly specialized at one thing, it may overfit and forget how to be good at everything else. 
% \item Difference to llm-guided evolution (AlphaEvolve / GEPA)\ZJ{We plan to compare ours  with Skydiscover}
% \end{itemize}
% }

Scientific discovery is inherently iterative. In domains such as protein engineering, progress depends not only on generating candidate solutions, but also on accumulating, revising, and reusing knowledge across repeated rounds of experimentation. Human experts rarely solve such tasks through isolated reasoning alone; instead, they progressively refine mechanistic heuristics from experimental feedback and reuse them in future design decisions.

Large language models (LLMs) possess strong reasoning capabilities and have recently shown promise in scientific domains including chemistry, biology, and materials discovery~\cite{song2025evaluatinglargelanguagemodels}. However, current LLM systems remain fundamentally limited in their ability to accumulate scientific expertise over time. Although they can generate candidate hypotheses or mutation proposals, they typically lack a persistent and structured mechanism for refining reusable knowledge across iterative experiments.

A major line of work adapts LLMs through reinforcement learning~\cite{ouyang2022training,guo2025deepseek}, encoding acquired knowledge in opaque model weights that are difficult to inspect or selectively revise. Without updating parameters, Self-Refine and Reflexion use feedback to refine outputs or guide subsequent trials~\cite{madaan2023self,shinn2023reflexion}, while AlphaEvolve~\cite{alphaevolve} and GEPA~\cite{agrawal2026gepa} evolve programs and prompts. These methods improve behavior through feedback, but leave open how to organize scientific experience into structured, reusable knowledge components that can be individually evaluated and revised across experiments.

% Existing approaches adapt LLM behavior primarily through parameter optimization, for example through reinforcement learning\cite{ouyang2022training,guo2025deepseek} . Without updating parameters, Self-Refine and Reflexion use feedback to refine outputs or guide subsequent trials~\cite{madaan2023self,shinn2023reflexion}.

% While effective in some settings, the acquired knowledge becomes entangled within opaque model weights, making it difficult to inspect, attribute, or selectively revise. Recent approaches, such as AlphaEvolve~\cite{alphaevolve} and prompt-evolution frameworks~\cite{agrawal2026gepa}, repeatedly revise generated artifacts based on evaluation feedback. However, these methods still treat learning as trajectory optimization: experience remains transient, localized to individual rollouts, and weakly structured for long-term reuse. As a result, current systems can improve outputs, but they do not explicitly accumulate reusable scientific expertise.

In this work, we argue that LLMs fall short in scientific reasoning not due to insufficient capacity, but because knowledge is not treated as a learnable object.
This observation motivates a shift in perspective: Instead of optimizing model parameters or solution trajectories, we propose to directly optimize externalized, persistent knowledge that conditions the model’s behavior. Under this view, learning is no longer about modifying the model itself, but about constructing and refining an explicit representation of domain knowledge.

However, optimizing knowledge directly introduces several fundamental challenges. First, knowledge does not naturally form a continuous optimization space: it is discrete, compositional, and structurally heterogeneous. Second, credit assignment becomes entangled because multiple knowledge components jointly influence each generated solution while feedback is often global. Third, evaluation signals are not directly usable for knowledge refinement, as scientific feedback is often sparse and scalar, providing limited guidance for structured updates.

To address these challenges, we propose \emph{KnowEvo}, a knowledge evolution framework for protein optimization. \emph{KnowEvo} is built on three key principles: (1) Transforming unstructured context into a structured and modular knowledge space; (2) Enabling credit assignment at the level of knowledge components; and (3) Converting evaluation signals into structured feedback for knowledge refinement. Together, these mechanisms make it possible to perform effective optimization in a discrete knowledge space. The main contributions of this paper are:

\begin{itemize}
\item We reformulate scientific discovery as optimization over externalized, persistent knowledge rather than over model parameters, decoupling domain adaptation from expensive finetuning.
\item We represent knowledge as executable expertise blocks organized into mutation trees, and propose a knowledge evolution mechanism that combines adaptive evaluation budget allocation with agentic evidence-driven revision.
\item We demonstrate that \emph{KnowEvo} outperforms root expert heuristics, monolithic program evolution, and prior LLM-guided evolution methods. %\XY{experiment results}
\end{itemize}

\section{Methodology}

\subsection{Overall Framework}
\textbf{Problem Formulation.}
We formulate protein optimization as an iterative optimization process over a knowledge base $K$, which is a collection of executable expertise blocks, each encoding a design mechanism as a concrete program. Let $\mathcal S$ denote the protein-state space, where a state $s\in \mathcal S$ contains the current protein sequence and any task-specific features used by the strategy blocks. Let $A(s)$ be the set of valid protein edit actions for state $s$, and let $\mathcal D$ denote the distribution of initial protein states.

Given a protein state \(s_t\), the current knowledge base $K$ selects a protein edit action \(a_{t}\sim q(\cdot\mid K, s_t)\) via sampling method $q$. The action then produces a new protein state $s_{t+1}=T(s_t,a_t)$ and a reward \[
r_t=R(s_t,s_{t+1})
=\mathbb{I}\!\left[\Delta(s_t,s_{t+1})>0\right],
\] where $T$ is the transition function and $\Delta(s_t,s_{t+1})$ denotes the task-specific improvement relative to the parent, such as a decrease in Rosetta energy or an increase in predicted solubility.
A rollout of length $H$ yields a trajectory \(\rho = (s_0, a_0, r_0, s_1, a_1, r_1, \ldots, s_H)\).
The objective is to maximize expected reward over initial states drawn from $\mathcal D$ and trajectories induced by $q$:
\begin{equation}\label{eq:obj}
  K^*=\arg\max_K J(K), \qquad
  J(K)=\mathbb{E}_{s_0\sim\mathcal D,\,\rho\sim q(\cdot\mid K,s_0)}
  \left[\sum_{t=0}^{H-1} r_t\right].
\end{equation}
Our experiments use single-step rollouts ($H=1$), with each candidate applied independently to its parent. With $H=1$, $J(K)$ is the probability of achieving a positive improvement.

\begin{figure}[t]
\centering
\includegraphics[width=\linewidth]{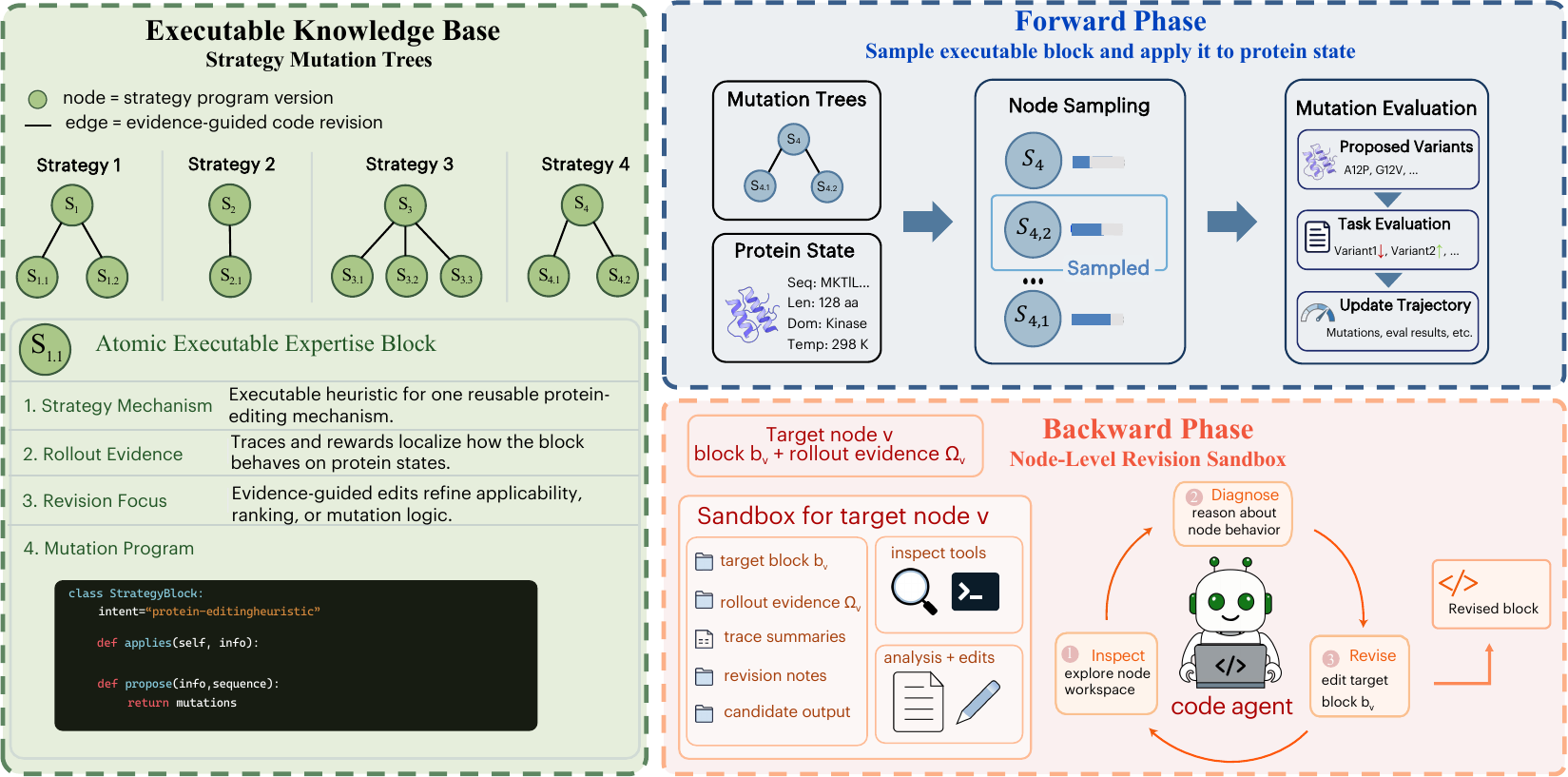}

\caption{\textbf{Overall framework of KnowEvo.}
KnowEvo evolves executable strategy mutation trees by alternating forward node evaluation with backward sandbox-based program revision.}
\label{fig:overall-framework}
\vspace{-0.8em}
\end{figure}

Optimizing a knowledge base introduces three distinct challenges:
(1) Representation: How to transform unstructured scientific heuristics into a structured form that supports fine-grained optimization and credit  attribution? 
(2) Feedback localization: How to convert sparse, global evaluation signals into structured, localized feedback for individual knowledge components?
(3) Context scaling: How to manage the large volume of execution trajectories without suffering from context rot, where critical information is lost or compressed beyond utility?
To address these, we introduce three corresponding components:
(1) {\it Executable expertise blocks} organized as {\it strategy mutation trees}, which offer a structured, externalized knowledge base for optimization;
(2) {\it Sampling algorithm}, which focuses on budget allocation and effective computational resource assignment;
(3) Code agent equipped with {\it agentic sandbox}, which maximizes the utilization of structured feedback from evaluation signals.
%Each of these components is elaborated on in the following sections.

% mutation trees and blocks
As shown in Figure~\ref{fig:overall-framework}, we organize the knowledge base as a mutation tree $\mathcal T_b=(\mathcal V_b,\mathcal E_b)$, where $\mathcal V_b$ and $\mathcal E_b$ denote the set of nodes and the set of edges in the tree. 
Each mutation tree $\mathcal T_b$ corresponds to a class of expert design mechanisms, and each strategy node $v\in\mathcal V_b$ stores a concrete block version, together with a set of trajectories $\{\rho\}_v$ from previous rollouts. 

\textbf{Training Process.}
Training proceeds through two asynchronous phases: the forward exploration phase and the backward evolution phase.
The {\it forward phase} evaluates current strategies against sampled protein states. Given a protein state $s_t \in \mathcal S$ at step $t$, the framework samples candidate strategies from the mutation trees $\{\mathcal T_b\}$ and executes each as an independent block-level rollout. During each rollout, the block proposes a protein edit action \(a_t\), which produces
\(s_{t+1}=T(s_t,a_t)\), and receives reward \(R(s_t,s_{t+1})\) on the task objective. The resulting trajectory set $\{\rho\}_v$ is then stored as structured information within the corresponding strategy nodes $v$.
In the {\it backward phase}, when a strategy node $v\in \mathcal V_b$ accumulates sufficient trajectories, i.e., ${\#\{\rho\}_v \ge N_\text{threshold}}\ $, a code agent revises its program within the sandbox; validated revisions are appended as child nodes in the same mutation tree $\mathcal T_b$.

\textbf{Budget-constrained expertise evolution.}
% \YM{Yanmming: Again, explain why budget-constrained}
Both node evaluation and strategy revision consume scarce resources.
Let $B_{\mathrm{eval}}$ be the total number of reward queries and $B_{\mathrm{work}}$ the agent-action allowance for each code revision; a separate cap limits the number of revision jobs.
The practical optimization problem reduces to a coupled budget allocation: given mutation trees $\{\mathcal{T}_b\}$ and budgets $B_{\mathrm{eval}}, B_{\mathrm{work}}$, allocate evaluation effort across nodes to estimate their utilities, and allocate revision effort to improve the most promising candidates, such that the expected deployment performance $J(K_{\mathrm{final}})$ is maximized.
The forward phase addresses evaluation allocation via Thompson-style posterior sampling; the backward phase addresses revision allocation through scalable agentic code refinement.

\subsection{Executable Expertise Blocks}
\label{sec:blocks}

% \ZJ{
% \begin{itemize}
%     \item  why using code to represent knowledge instead of language like prompt? 
% deterministic, more structured and is able to compute-> concrete. easier to attribute
% \item why designing atomic code block
% \item how to design such block 
% \end{itemize}
% }
% \YM{When introduce this section, it is better to describe it with a figure, either with the overall system figure (with blocks) or a separate ru}

% \XY{First introduce why we use blocks (structured knowledge. Secondly, why use executable blocks instead of natural language blocks? Finally, explain what the block exactly is.}

A central challenge in applying large language models to scientific discovery lies in transforming unstructured textual knowledge into a representation that allows better systematic optimization and resource assignment. %Rather than encoding knowledge implicitly in model parameters which is expensive and domain specialized, we externalize the knowledge as persistent and reusable blocks, which shifts the optimization target from model weights to the knowledge artifacts themselves.
In this work, we adopt executable strategy blocks instead of natural-language prompts as the form of externalized knowledge.
Although natural language can describe scientific heuristics, it lacks explicit operational semantics: applying a heuristic to a concrete protein state requires the LLM to re-interpret the prompt at every invocation, making execution non-deterministic and evaluation costly.
We instead concretize each piece of knowledge as an executable program with explicit inputs, parameters, and mutation outputs, making it deterministic and revisable at a mechanism-level granularity.

Each executable expertise block can be viewed as a mechanism-level scientific hypothesis: it proposes that a particular design mechanism is useful under a particular set of conditions. Rather than encoding the entire protein-design policy in one monolithic program, a block isolates one strategy component with explicit applicability logic and mutation behavior. This decomposition aligns the unit of execution with the unit of evidence. When a block succeeds or fails, the resulting trajectory provides a localized signal about that mechanism, rather than an entangled signal about a full policy. The system can therefore revise the block's applicability boundary, action logic, or internal parameters without overwriting unrelated knowledge.

We therefore organize the executable knowledge base as a collection of strategy mutation trees. For each root expertise block \(b\in\mathcal B\), the framework
maintains a tree \(\mathcal T_b=(\mathcal V_b,\mathcal E_b)\), where each node \(v\in\mathcal V_b\) is an executable version of the same mechanism-level strategy, and each edge \((u,v)\in\mathcal E_b\) records an accepted evidence-driven revision from parent \(u\) to child \(v\). Thus, the knowledge base \(K=\{\mathcal T_b\}_{b\in\mathcal B}\) preserves the revision lineage of each mechanism while exposing all validated versions as candidate strategies for future execution.

\subsection{Evaluation Budget Allocation in Forward Phase}
\label{sec:budget-allocation-forward-phase}

In the forward phase, the evaluation budget is allocated across candidate nodes in the mutation trees. Each node $v \in \mathcal{V}_b$ is a complete, independently executable expertise block. Its utility $U_v$ is unknown a priori and needs to be inferred from execution outcomes. 
As the mutation tree grows, the system continuously decides which nodes to evaluate, balancing exploitation of reliable nodes against exploration of uncertain or recently revised ones. We therefore cast forward budget allocation as an \emph{online posterior allocation} problem over a dynamically expanding candidate node set $\mathcal V_b$. 
This formulation differs from tree search methods such as Monte Carlo Tree Search~(MCTS)~\cite{coulom2006efficient}, where nodes represent intermediate states and the objective is path selection rather than allocation over independent alternatives.

% <ver2>
\paragraph{Evaluation budget as online posterior allocation.}
We model the outcome of executing a strategy node \(v\) on a given input as a
binary success/failure with unknown success probability \(p_v\). The utility
\(U_v\) is identified with \(p_v\). 

Let \(\mathcal H_v(t)=(S_v(t),F_v(t))\) be the cumulative numbers of successes
and failures observed for node \(v\) before the \(t\)-th forward rollout. The
posterior distribution of \(p_v\) takes the following form:
\begin{equation}
p(p_v \mid \mathcal H_v(t)) = \mathrm{Beta}\big(S_v(t)+1,\; F_v(t)+1\big).
\end{equation}

Specifically, nodes without any observation inherit an uninformative prior \(\mathrm{Beta}(1,1)\), which is uniform on \([0,1]\). 
For each incoming task, the framework forms an \emph{eligible node set}
\(\mathcal A_b(t)\subseteq\mathcal V_b\) for the mutation tree and draws a
posterior sample of the success probability for every candidate:
\begin{equation}
  \tilde p_v(t) \sim \mathrm{Beta}\big(S_v(t)+1,\; F_v(t)+1\big),
  \qquad v\in\mathcal A_b(t).
\end{equation}
These samples are converted into a probability distribution over nodes via a
softmax with a fixed temperature \(\tau>0\):
\begin{equation}
  \pi_{b,t}(v)
  =
  \frac{
    \exp(\tilde p_v(t)/\tau)
  }{
    \sum_{u\in\mathcal A_b(t)} \exp(\tilde p_u(t)/\tau)
  },
  \qquad v\in\mathcal A_b(t).
\end{equation}
When \(\tau\to 0\), the softmax reduces to the conventional Thompson sampling,
concentrating on the node with the largest sampled \(\tilde p_v\). In this
limit, the selection probability matches the posterior probability that a node
is optimal:
\begin{equation}
  \Pr(v_t = v \mid \mathcal H_t) =
  \Pr\!\Big(v = \arg\max\nolimits_{u\in\mathcal A_b(t)} p_u \;\big|\; \mathcal H_t\Big),
\end{equation}
under the independent Beta--Bernoulli model. Here $\mathcal H_t$ denotes all node-outcome histories available before rollout $t$.
Our growing mutation trees use a fixed positive temperature to retain exploration of eligible revisions. This is a softmax variant of Thompson sampling; we do not claim a stationary-bandit regret guarantee for this adaptive setting.
% simplify the statement?

\medskip
\textbf{Discovery of new nodes.}
Every eligible node receives positive selection probability. If at most $M$
nodes are eligible in a rollout, then $0\leq\tilde p_u\leq1$ implies
\begin{equation}
  \pi_{b,t}(v)
  =\left[1+\sum_{u\ne v}\exp\!\left(\frac{\tilde p_u-\tilde p_v}{\tau}\right)\right]^{-1}
  \geq \frac{1}{1+(M-1)e^{1/\tau}}=: \delta(\tau,M).
\end{equation}
If a new node remains eligible for $h$ consecutive rollouts and the eligible
set never exceeds $M$, multiplying the conditional probabilities of avoiding
it gives a probability of at least $1-(1-\delta)^h$ of evaluating it once.
The bound improves with $\tau$ and weakens with $M$; it does not guarantee
rapid evaluation when the temperature is small or many nodes compete.

\medskip
\textbf{Allocation dynamics.}
For a total evaluation budget \(B_{\mathrm{eval}}\), the expected number of
rollouts assigned to node \(v\) is
\begin{equation}
  \mathbb E\big[N_v(B_{\mathrm{eval}})\big]
  =
  \sum_{t=1}^{B_{\mathrm{eval}}}
  \mathbb E\big[\pi_{b,t}(v)\big].
\end{equation}
Here \(N_v(B_{\mathrm{eval}})\) denotes the number of rollouts assigned to node \(v\).
Here $\pi_{b,t}(v)=0$ whenever $v$ is not yet created or is ineligible.
At fixed $\tau>0$, softmax continues to assign positive mass to eligible
alternatives even if their posteriors concentrate. This residual exploration
comes from the temperature, rather than persistent posterior uncertainty.
Allocation is updated after each rollout, allowing newly revised nodes to
compete with established strategies as evidence accumulates.
% </ver2>

\subsection{Agentic Revision Sandbox}
\label{sec:agentic-revision-sandbox}

Scientific optimization requires turning evaluation feedback into an improved
executable strategy. In our setting, each strategy node is a program. A direct
revision operator could serialize the accumulated feedback into an LLM prompt 
and ask for a revised program, but this reduces revision to a single-step text
generation process. Given the substantial and heterogeneous feedback in our framework, such a prompt-only operation is insufficient and may suffer from context rot.

We therefore perform backward evolution inside a \emph{revision sandbox}. The
sandbox is an isolated executable environment that materializes the target
block, rollout evidence, runtime traces, task schemas, analysis scripts, and
validation tests as persistent external state. It is distinct from the task
environment \(\mathcal E\), which evaluates candidate mutations. The code
agent operates inside the revision sandbox: it can inspect evidence on demand,
run analyses, edit the target block, execute validation scripts, and repair
runtime failures before submitting a successor strategy. % \YM{Don't need to mention specific agents. The agentic revision sandbox section is kind of implemental, we need to emphasize the technical things. Otherwise, the reviewers will think we just apply CC. We need to explain what we do on top of CC}

For a strategy node \(v\) with executable block \(b_v\), the forward phase
spends an evaluation budget \(N_v(B_{\mathrm{eval}})\) in the task environment \(\mathcal E\),
producing an observation
\begin{equation}
    \Omega_v^{(N_v(B_{\mathrm{eval}}))} \sim \mathrm{Observe}_{N_v(B_{\mathrm{eval}})}(b_v;\mathcal E).
\end{equation}
The observation \(\Omega_v^{(N_v(B_{\mathrm{eval}}))}\) is then materialized into the revision
sandbox rather than compressed into a flat prompt. The agent receives bounded
views of this external state and acts through tool calls to analyze evidence,
modify the program, and validate the revised block.

We model the backward phase as an agentic revision operator with internal work
budget \(B_\text{work}\):
\begin{equation}
    b_{v'} \sim
    \mathcal A_{\phi,B_\text{work}}(b_v,\Omega_v^{(N_v(B_{\mathrm{eval}}))},J).
\end{equation}
Here \(J\) denotes the task objective, and \(B_\text{work}\) controls how many
agent actions can be spent on inspection, analysis, editing, execution, and
repair. A proposed successor is accepted only if it satisfies the runtime
contract and passes validation. An accepted proposal is inserted into the same
strategy mutation tree as a child node, where it can receive future evaluation
budget through subsequent rollouts.

This abstraction separates evidence scaling from agentic-work scaling. The
forward budget \(B_\text{eval}\) controls how much experimental evidence is generated, the
retained sandbox state controls how much evidence remains available outside the
agent's immediate context window, and \(B_\text{work}\) controls how much internal work
can be spent converting that evidence into a validated successor program. Thus,
large scientific feedback need not be compressed into a single prompt; it
becomes an executable, inspectable, and revisable external state.

\begingroup

\paragraph{Inference selection and candidate ranking.}
After training, we select one node per strategy family by maximizing the lower confidence bound (LCB), defined as the 5th percentile of its Beta posterior estimated from training rollouts.
Each selected node proposes one mutation set, applied independently to the parent protein to produce one candidate sequence.
We rank these candidates by their nodes' training-time LCB scores in
descending order (Appendix~\ref{app:inference-selection-lcb}).
\par
\endgroup

% Compact spacing between experimental floats and text.
\begingroup
% LAYOUT REPAIR (2026-11-01): regular, compact gaps around experimental floats.
\setlength{\textfloatsep}{8pt plus 2pt minus 2pt}
\setlength{\floatsep}{6pt plus 2pt minus 2pt}
\setlength{\intextsep}{6pt plus 2pt minus 2pt}
\section{Experiments}
\label{sec:experiments}

%\YM{List the experiments to be included.
%(1) Overall Performance for various tasks (Baselines include vanilla LLM, RL, GEPA, Skydiscover?). (2) Ablations (block, , Thompson sampling?) (3) If the learned mutations trees do indicate some explainable knowledge, we may show a case study. Or at least we need to show the mutation trees before/after update}

In the experiments, we try to answer the following questions:
\textbf{RQ1: Does knowledge evolution improve task performance?} % main results
We compare revised strategy trees against their root seed blocks and mutation-generation baselines under matched reward-call budgets.
% \textbf{RQ2: How do the core design choices affect performance?}
% We ablate the forward allocation policy, the revision sandbox, and the mutation-tree structure.
\textbf{RQ2: How do the core designs work and affect performance?}
We trace the revision lineage of strategy blocks to examine how rollout evidence drives localized changes in design mechanisms and progressively refines executable expertise.
We ablate the forward allocation policy, the revision sandbox, and the mutation-tree structure.
\textbf{RQ3: Do the gains from knowledge evolution generalize across evaluation predictors?}
We evaluate Rosetta-evolved strategies using an independent stability predictor (ThermoMPNN) to assess whether improvements transfer beyond the optimization backend. 
\textbf{RQ4: How does strategy evolution scale with evidence and agentic work?}
We vary the number of trajectories available for revision and the code agent's internal work budget to separate information scaling from agentic-work scaling.
% RQ4: How does executable knowledge evolve through iterative revisions? We trace the revision lineage of strategy blocks to examine how rollout evidence drives localized changes in design mechanisms and progressively refines executable expertise.

\subsection{Experimental Settings}

\paragraph{Tasks.} We evaluate on two BRENDA-derived protein optimization tasks: stability, scored by Rosetta REF2015 energy, and solubility, scored by a SaProt-based solubility predictor. For each task, we use 5000 training proteins and a 200-protein held-out split. After backend-validity filtering, evaluation uses 198 proteins for stability and 196 for solubility; all methods within each task use the same subset.

% \paragraph{Protocol.} A candidate is counted as successful if it improves the active backend score relative to the parent sequence. For evaluation, every mutation generator is exposed through a common ranked candidate-list interface. Given a parent protein, a method returns \(\mathcal C=(M_1,\ldots,M_m)\), \(m\leq 5\), where each \(M_i\) is a mutation set applied independently to the parent sequence, and the index \(i\) defines its proposed rank. This includes the monolithic-program baseline, whose program is required to emit a ranked list rather than a single mutation. We report pass@\(k\), which measures whether any successful candidate appears within the top-\(k\) list, together with MRR, MAP, and NDCG@5 to measure whether successful candidates are ranked early. Unless otherwise stated, all backward revisions in the main experiments and ablation studies use MiniMax-M2.7\cite{minimax2026m27} API to power code agent. 

\paragraph{Protocol.}
For evaluation, every mutation generator is exposed through a common ranked candidate-list interface. Given a parent protein, a method returns \(\mathcal C=(M_1,\ldots,M_m)\), \(m\leq 5\), where each \(M_i\) is a mutation set applied independently to the parent sequence, and the index \(i\) defines its proposed rank. This includes the monolithic-program baseline, whose program is required to emit a ranked list rather than a single mutation. Unless otherwise stated, all backward revisions in the main experiments and ablation studies use the MiniMax-M2.7~\cite{minimax2026m27} API to power the code agent.

\paragraph{Metrics.}
A candidate is counted as successful if it improves the active backend score relative to the parent sequence. We report pass@\(k\), which measures whether any successful candidate appears within the top-\(k\) list, together with MRR, MAP, and NDCG@5 to evaluate how effectively each method ranks successful candidates toward the top of the list.

\paragraph{Baselines.} We compare against methods that optimize different objects: root seed blocks,
which evaluate the initial executable heuristics without revision; a monolithic
program baseline, which flattens the seed heuristics into a single mutation
generator but uses the same ranked candidate-list interface; and
% BEGIN JOINT BASELINE INTRODUCTION (2026-11-01): blue marks changed content.
AdaEvolve~\cite{cemri2026adaevolve} and
GEPA~\cite{agrawal2026gepa}, two LLM-guided evolution baselines evaluated
under the same reward-call budget.
We additionally evaluate \emph{Agentic AdaEvolve}, which uses the same sandboxed code agent as KnowEvo for iterative program revision; LLM token usage is reported separately in Appendix~\ref{app:llm-cost}.
% END JOINT BASELINE INTRODUCTION
When available, we also include task-specific policy optimization baselines.
% Detailed baseline instantiations are provided in Appendix~\ref{app:baselines}.

Full experimental settings are given in Appendix~\ref{app:main-exp-details}.
% Full dataset construction, backend configurations, validity statistics, detailed baseline instantiations, and metrics are given in Appendix~\ref{app:main-exp-details}.

 \subsection{Main Results}

Table~\ref{tab:main_results} shows that knowledge evolution consistently
improves performance across both tasks. On Rosetta stability, our
method improves over root seed blocks from \(42.4\%\) to \(50.5\%\) in pass@1
and from \(74.2\%\) to \(82.8\%\) in pass@3. On SaProt solubility, the gains are
larger: pass@1 increases from \(31.6\%\) to \(51.5\%\), and pass@3 increases
from \(57.7\%\) to \(83.2\%\). The same trend holds for MRR, MAP, and NDCG@5,
indicating that successful mutations are not only found more often but also
ranked earlier.
% ranked earlier. Qualitative case studies in Appendix~\ref{app:case-studies} illustrate how rollout evidence is converted into executable residue-level rules.

KnowEvo outperforms GEPA and both
AdaEvolve variants across all six metrics on both tasks. Against
Agentic AdaEvolve, which uses the same sandboxed revision agent,
KnowEvo improves MRR by 6.2 points on stability and 9.8 points
on solubility.

% BEGIN RESULTS REFERENCES ADDITION (2026-11-01): blue marks new content.
{Appendix~\ref{app:llm-cost} presents a detailed analysis on LLM token usage and performance.}

\begin{table}[!htbp]
\caption{Comparisons across stability and solubility tasks. Bold indicates the best result for each task.}
\label{tab:main_results}

\centering
\small
\setlength{\tabcolsep}{4.8pt}
\renewcommand{\arraystretch}{1.05}

\newcommand{\best}[1]{\textbf{#1}}

\begin{tabular*}{\linewidth}{@{\extracolsep{\fill}}lcccccc@{}}
\toprule
Method
& \multicolumn{3}{c}{Hit rate}
& \multicolumn{3}{c}{Ranking quality} \\
\cmidrule(lr){2-4}
\cmidrule(l){5-7}
& pass@1 $\uparrow$ & pass@3 $\uparrow$ & pass@5 $\uparrow$
& MRR $\uparrow$ & MAP $\uparrow$ & NDCG@5 $\uparrow$ \\
\midrule

\rowcolor{gray!10}
\multicolumn{7}{@{}l}{\textit{Stability / Rosetta}} \\
Direct LLM proposal
& 37.9 & 70.2 & 82.3 & 54.7 & 54.7 & 61.7 \\
Pro-1 8B
& 42.4 & 69.2 & 77.3 & 56.4 & 53.2 & 61.0 \\

\addlinespace[2pt]
Monolithic program
& 43.4 & 72.2 & 83.8 & 58.7 & 55.6 & 64.2 \\
GEPA
& 37.9 & 70.7 & 83.8
& 55.0 & 54.1 & 62.7\\
AdaEvolve
& 44.4 & 74.7 & 84.8
& 60.5 & 58.5 & 66.5
\\
\quad + agentic revision
& 45.5 & 77.8 & 81.8
& 60.2 & 59.1 & 65.5
\\

\addlinespace[2pt]
Root seed blocks
& 42.4 & 74.2 & 85.4 & 59.1 & 56.3 & 65.2 \\
KnowEvo (Ours)
& \best{50.5} & \best{82.8} & \best{88.4}
& \best{66.4} & \best{63.2} & \best{71.2} \\

\midrule

\rowcolor{gray!10}
\multicolumn{7}{@{}l}{\textit{Solubility / SaProt}} \\
Direct LLM proposal
& 50.5 & 61.7 & 63.8 & 56.3 & 56.3 & 58.2 \\

\addlinespace[2pt]
Monolithic program
& 35.2 & 65.8 & 79.6 & 51.9 & 49.2 & 58.3 \\
GEPA
& 40.3 & 76.0 & 87.2
& 59.4 & 58.2 & 66.7

\\
AdaEvolve
& 49.0 & 74.0 & 82.5 & 61.9 & 59.8 & 66.9 \\
\quad + agentic revision
& 42.3 & 79.6 & 80.6
& 58.0 & 58.1 & 64.2
 \\
\addlinespace[2pt]
Root seed blocks
& 31.6 & 57.7 & 85.2 & 47.2 & 47.5 & 57.5 \\
KnowEvo (Ours)
& \best{51.5} & \best{83.2} & \best{90.8}
& \best{67.8} & \best{64.2} & \best{72.8} \\

\bottomrule
\end{tabular*}
\end{table}

% BEGIN MECHANISM EVOLUTION ADDITION (2026-10-29): blue marks new content.
% BEGIN MECHANISM EVOLUTION ADDITION (2026-10-29)
\FloatBarrier
\begingroup
\subsection{Analysis of Revision Process}
\label{sec:mechanism-evolution}
To examine how executable knowledge changes across successive revisions,
we trace the solubility multi-site strategy from one complete evolution path in Figure~\ref{fig:mechanism-evolution} (the details of the entire evolution path are provided in
Appendix~\ref{app:solubility-evolution-path}). The strategy progresses
from combining hydrophobic, charge, and aromatic edits to conditioning
substitutions on global composition, local context, and sequence position.

\begin{figure}[!htbp]
\centering
\includegraphics[width=\linewidth]{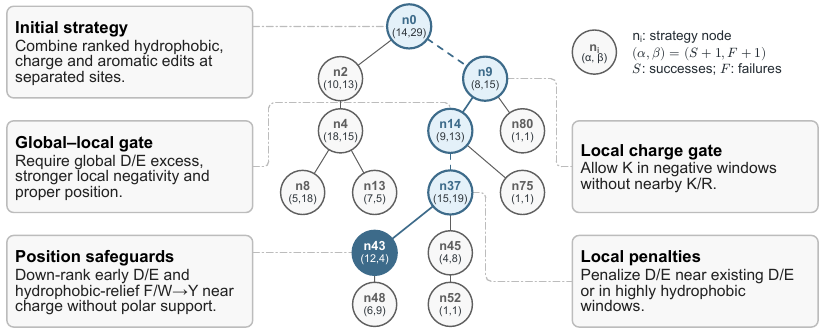}
\caption{Evolution of the solubility strategy.
Callouts highlight changes to executable rules along the lineage; dashed arrows omit intermediate nodes. Node pairs are Beta
posterior parameters \((\alpha,\beta)=(S+1,F+1)\), where \(S\) and \(F\)
count successful and unsuccessful evaluations at that node.}
\label{fig:mechanism-evolution}
\end{figure}

\begin{samepage}
 For example, evolution from node 9 to node 14 refines local charge filtering into a policy that coordinates local sequence context with protein-wide composition. Building on restrictions near basic residues, node 14 allows positive-charge edits only when both the local environment and the overall protein favor acidic residues, while adding protection for the N-terminal region. 

\end{samepage}

% BEGIN CASE STUDY REFERENCE ADDITION (2026-11-01): blue marks new content.
 Appendix~\ref{app:case-studies}
presents a more detailed analysis of how feedback leads to mechanism evolution for another two tasks.
% END CASE STUDY REFERENCE ADDITION

% LAYOUT REPAIR (2026-11-01): close the analysis scope and contain its figure.
\par
\endgroup
\FloatBarrier
% END MECHANISM EVOLUTION ADDITION

% END MECHANISM EVOLUTION ADDITION

\subsection{Ablation Study and Sensitivity Analysis}
\label{sec:rq2-ablation}

We conduct a diagnostic ablation study on the solubility/SaProt task to isolate
the contribution of each core design choice in the knowledge-evolution loop.
We use solubility for this analysis because its reward backend is substantially
cheaper than Rosetta and supports repeated controlled runs under matched
settings. 

We ablate three components. First, we replace the Thompson-style posterior
allocation with uniform random node sampling, testing whether adaptive
evaluation over strategy nodes is useful. Second, we remove the revision sandbox
and ask a single LLM call to directly revise the target block from serialized
evidence, testing whether persistent external state and tool-based repair are
needed. Third, we replace the mutation tree with chain-only growth, where each
strategy family keeps only the latest accepted successor, testing whether
retaining competing ancestors and siblings improves final strategy selection.

\begin{table}[!htbp]
\caption{Ablation study on the solubility/SaProt task. All variants use the same reward-call budget and held-out candidate-list evaluation protocol.}
\label{tab:rq2_ablation}

\centering
\small
\setlength{\tabcolsep}{5.2pt}
\renewcommand{\arraystretch}{1.05}

\newcommand{\best}[1]{\textbf{#1}}

\begin{tabular*}{\linewidth}{@{\extracolsep{\fill}}lcccccc@{}}
\toprule
Method
& \multicolumn{3}{c}{Hit rate}
& \multicolumn{3}{c}{Ranking quality} \\
\cmidrule(lr){2-4}
\cmidrule(l){5-7}
& pass@1 $\uparrow$ & pass@3 $\uparrow$ & pass@5 $\uparrow$
& MRR $\uparrow$ & MAP $\uparrow$ & NDCG@5 $\uparrow$ \\
\midrule

Random node allocation
& 51.0 & 78.5 & 88.2
& 65.4 & 61.8 & 69.9 \\

Single-step LLM revision
& 49.0 & 80.1 & 84.2
& 64.2 & 62.6 & 69.3 \\

Chain-only revision
& 43.4 & 54.1 & 54.1
& 48.7 & 48.7 & 50.1 \\

KnowEvo (Full method)
& \best{51.5} & \best{83.2} & \best{90.8}
& \best{67.8} & \best{64.2} & \best{72.8} \\

\bottomrule
\end{tabular*}
\end{table}

% \XY{Table~\ref{tab:rq2_ablation} shows that each component contributes to the final
% performance.
% Random node allocation shows improvement over root blocks and weakness comparing to full method.
% Random node allocation still improves over the root seed blocks, because revision remains active and the final deployed nodes are selected by the same LCB-based rule as in the full method. Thus, random allocation can still produce useful nodes. Its weakness is inefficient budget use: promising nodes are not preferentially evaluated and therefore are not fully exploited. 
% The single-step LLM revision variant also improves over weaker baselines but is consistently worse than the full method, suggesting that serializing evidence into one prompt is less reliable than preserving details as external sandbox state.
% Chain-only revision performs worst, showing that linear revision is insufficient. 
% Overall, these ablations support the coupled design of our framework: revision creates useful successors, posterior allocation makes evaluation budget responsive to evidence, and the mutation tree preserves alternatives for robust final LCB-based selection.}

% Observations -> Reasons
Table~\ref{tab:rq2_ablation} shows the contribution of each component to the overall performance.
Random node allocation outperforms the root seed blocks but underperforms the full method: revision still produces useful successors selected by the same LCB rule, while uniform allocation spends less evaluation effort on promising nodes. Single-step LLM revision also underperforms the full method, consistent with a benefit from retaining detailed evidence in the sandbox. Chain-only revision performs worst, consistent with the loss of alternatives when ancestors are discarded.
Overall, these ablations support the coupled design of our framework: revision creates useful successors, posterior allocation makes evaluation budget responsive to evidence, and the mutation tree preserves alternatives for robust final LCB-based selection.

We additionally evaluate three alternatives under the same solubility/SaProt backend  as the human-written initialization.
\emph{Cold start} uses executable roots that propose random mutations.
\emph{Warm start} also uses randomized executable roots, but assigns
each block a fixed exploration direction corresponding to the
mechanism identities in the human-written initialization.
\emph{LLM-generated} initialization uses five roots generated
sequentially by MiniMax-M2.7 to reduce overlap between mechanisms.
Table~\ref{tab:initialization-sensitivity} reports final performance
after evolution from each initialization.

% LAYOUT REPAIR (2026-11-01): avoid flushing this table onto a mostly empty page.
\begin{table}[H]
\caption{
Sensitivity to initial knowledge on solubility/SaProt. Values are percentages; bold indicates the highest value
for each metric.
}
\label{tab:initialization-sensitivity}
\centering
\small
\setlength{\tabcolsep}{4pt}
\renewcommand{\arraystretch}{1.05}
\begin{tabular*}{\linewidth}{@{\extracolsep{\fill}}lcccccc@{}}
\toprule
Initialization
& \multicolumn{3}{c}{Hit rate}
& \multicolumn{3}{c}{Ranking quality} \\
\cmidrule(lr){2-4}
\cmidrule(l){5-7}
& pass@1 $\uparrow$ & pass@3 $\uparrow$ & pass@5 $\uparrow$
& MRR $\uparrow$ & MAP $\uparrow$ & NDCG@5 $\uparrow$ \\
\midrule
Human-written
& 51.5 & 83.2 & 90.8 & 67.8 & 64.2 & 72.8 \\
Cold start
& 46.4 & 79.1 & \textbf{93.9} & 64.2 & 60.1 & 70.6 \\
Warm start
& \textbf{54.6} & 81.1 & 91.8 & \textbf{69.3} & 62.6 & 72.4 \\
LLM-generated
& 51.5 & \textbf{85.7} & 92.4 & 68.7 & \textbf{66.8} & \textbf{74.5} \\
\bottomrule
\end{tabular*}
\end{table}

Table~\ref{tab:initialization-sensitivity} shows that KnowEvo achieves competitive performance across different initializations. These results suggest that KnowEvo is robust to root-block initialization in this setting and does not critically depend on manually written expert mutation rules.

% BEGIN CROSS-EVALUATION ADDITION (2026-10-28): blue marks the added content.

\FloatBarrier
\subsection{Cross-Predictor Evaluation}
\label{sec:cross-predictor-evaluation}
We assess whether gains from Rosetta-guided evolution carry over to a second
stability predictor. We re-score the candidates generated by the initial root
strategies and the final KnowEvo strategies with
ThermoMPNN~\cite{dieckhaus2024thermompnn}, which is used only for this post-hoc
evaluation, with no additional strategy evolution.
Table~\ref{tab:cross-predictor-main} reports higher point estimates for KnowEvo
across all six metrics, providing evidence that  gains extend beyond the
optimization backend. Details are provided in Appendix~\ref{app:cross-predictor}.

\begin{table}[!htbp]
\centering
\small

\caption{Cross-predictor evaluation on stability.
ThermoMPNN re-scores candidates from initial and Rosetta-evolved strategies
and is not used for evolution. Values are point estimates multiplied by 100.}
\label{tab:cross-predictor-main}
\renewcommand{\arraystretch}{1.05}
\begin{tabular*}{\linewidth}{@{\extracolsep{\fill}}lrrrrrr@{}}
\toprule
Strategy & pass@1 & pass@3 & pass@5 & MRR & MAP & NDCG@5 \\
\midrule
Initial & 9.1 & 31.3 & 36.9 & 20.5 & 20.5 & 24.7 \\
\textbf{KnowEvo} & \textbf{15.7} & \textbf{32.8} & \textbf{37.4}
& \textbf{23.7} & \textbf{23.1} & \textbf{26.9} \\
\bottomrule
\end{tabular*}
\end{table}
\FloatBarrier
% END CROSS-EVALUATION ADDITION

\subsection{Scaling External Knowledge Through Agentic Revision}
\label{sec:exp-agentic-revision-scaling}

RQ4 asks whether the revision sandbox can convert additional resources into
better executable strategy knowledge. We consider two levels: extending the evolutionary process as rollout evidence accumulates, and allocating more internal agentic work to each revision. We also examine how the revision backbone affects this process. All analyses use solubility/SaProt, which supports frequent checkpoint evaluation.
% First,
% as more forward rollouts are accumulated, the sandbox receives more trajectory evidence for revision; in our implementation this is reflected by more revisions in the strategy tree. 
%  First, We track the strategies discovered over successive revisions
% Second, when the evidence input is fixed, the
% code agent may spend more internal work converting that evidence into program
% changes. Third, under the same evidence input and revision protocol, a stronger
% code agent may perform this evidence-to-code conversion more effectively. We use
% solubility/SaProt as a dense diagnostic setting because it supports frequent
% checkpoint evaluation.

\paragraph{Scaling the evolution budget.}
We first examine whether continued evolution discovers better strategies. We run six independent
revision trajectories and evaluate the strategy tree every ten revisions on the same frozen diagnostic split. For each trajectory, we report
the best-so-far performance available up to that point, and then average across trajectories.

\begin{figure}[!htbp]
\centering
\includegraphics[width=0.90\linewidth]{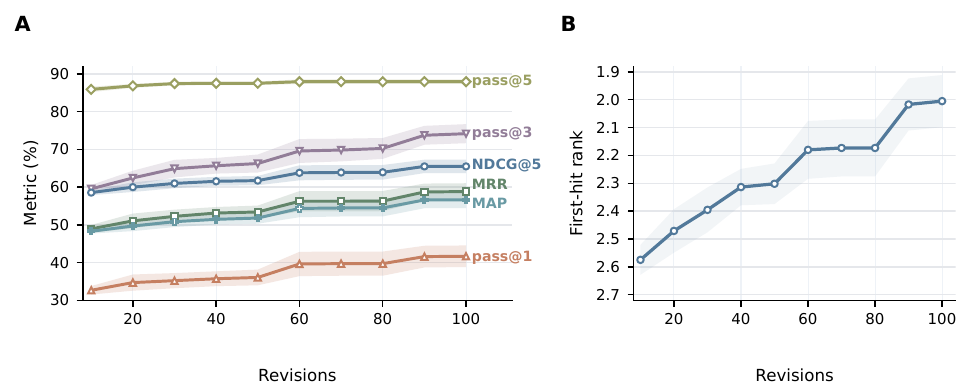}
\caption{Evolution budget scaling through accumulated revisions on solubility/SaProt.
Best-so-far checkpoint performance is averaged over six revision trajectories.
(A) Ranking metrics across revisions. (B) Average first-hit rank: lower is better.}
\label{fig:rq3-agentic-revision-scaling}
\end{figure}

\begin{samepage}
Figure~\ref{fig:rq3-agentic-revision-scaling} summarizes the best-so-far diagnostic performance as evolution proceeds. From 10 to 100 accepted
revisions, pass@3 rises from \(59.5\%\) to \(74.2\%\), while the average
first-hit rank decreases from \(2.58\) to \(2.01\). The pass@5 score is already high and
changes only modestly, but early-rank and ranking-sensitive metrics such as
pass@1, MRR, MAP, and NDCG@5 improve. Because these curves retain the best observed checkpoint, they are non-decreasing by construction and do not imply that every revision improves performance.
\par
\end{samepage}

\paragraph{Scaling the revision work.}
We next examine whether additional work within each revision improves  performance. We compare runs under the same 100-revision budget and evidence-collection protocol, while varying the internal revision-loop budget ($\kappa$). Larger budgets allow additional diagnostic passes before the agent produces the revised block. We record cumulative agent turns as a proxy for realized effort. The protocol is detailed in Appendix~\ref{app:agentic-revision-loop-details}.

\begin{figure}[!htbp]
\centering
\includegraphics[width=0.90\linewidth]{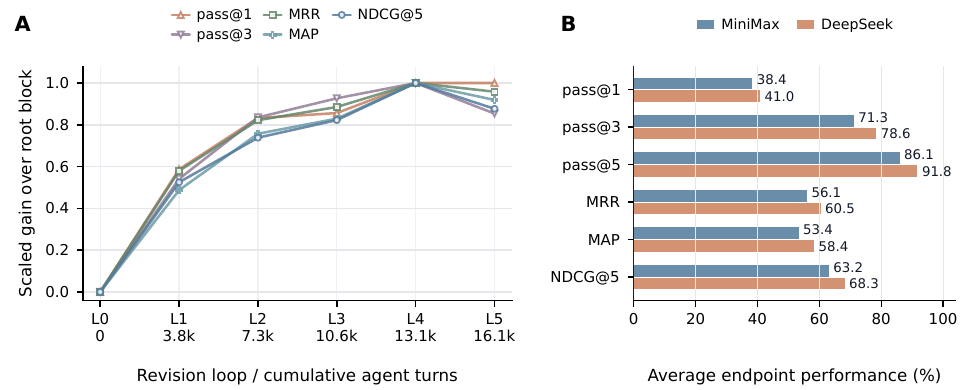}
\caption{Scaling the evidence-to-code conversion on solubility/SaProt.
(A) Agent-work scaling under the same revision-count budget: realized code-agent work is measured by cumulative agent turns, and the
y-axis reports scaled endpoint gains over the initial root-block selection. (B) Revision-backbone diagnostic under the same revision-count budget. }
\label{fig:rq3-agentic-conversion}
\end{figure}

Figure~\ref{fig:rq3-agentic-conversion}A shows higher endpoint gains with more sandbox work under the same revision-count budget and evidence-collection protocol. This supports additional revision effort, but does not isolate work on identical evidence packets, since revised strategies can change subsequent rollouts. 
\paragraph{Effect of the revision backbone.}
Finally, we compare MiniMax and DeepSeek under the same  100-revision budget and evaluation protocol. Figure~\ref{fig:rq3-agentic-conversion}B shows that the tested DeepSeek configuration achieves higher average endpoint performance across all reported metrics. This result suggests that the revision backbone provides another avenue for improving the effectiveness of knowledge evolution.

\section{Conclusion}
We presented \emph{KnowEvo}, a framework that recasts LLM-based protein optimization as the iterative refinement of externalized, persistent knowledge rather than model parameters or transient solution trajectories. By combining executable expertise
blocks, strategy mutation trees, and a Thompson-sampling forward allocator with an agentic revision sandbox in the backward phase, \emph{KnowEvo} converts sparse scalar rewards into localized, inspectable program revisions. Experiments on Rosetta stability and SaProt solubility demonstrate that \emph{KnowEvo} achieves higher performance than root seed blocks, monolithic program evolution, and LLM-guided evolution baselines across pass@$k$, MRR, MAP, and
NDCG@5. Our scaling study further shows that the gains compound with more rollout evidence and stronger code agents. These results
indicate that making knowledge itself the optimization target is a practical
path toward LLM systems that accumulate, attribute, and reuse scientific expertise over time.

% LAYOUT REPAIR: flush experimental floats before restoring global spacing.
\par
\clearpage
\endgroup
\begin{ack}
This work is supported in part by the National Natural Science Foundation of China under Grant
62276044, and also in part by the 2025 Scientific
Research Projects of the General Administration
of Customs of China under Grant No. 2025HK184.
\end{ack}

% References section
% \section*{References}
{
\small
\bibliographystyle{unsrt}
\bibliography{references}
}

% References follow the acknowledgments in the camera-ready paper. Use unnumbered first-level heading for
% the references. Any choice of citation style is acceptable as long as you are
% consistent. It is permissible to reduce the font size to \verb+small+ (9 point)
% when listing the references.
% Note that the Reference section does not count towards the page limit.
% \medskip

% {
% \small

% [1] Alexander, J.A.\ \& Mozer, M.C.\ (1995) Template-based algorithms for
% connectionist rule extraction. In G.\ Tesauro, D.S.\ Touretzky and T.K.\ Leen
% (eds.), {\it Advances in Neural Information Processing Systems 7},
% pp.\ 609--616. Cambridge, MA: MIT Press.

% [2] Bower, J.M.\ \& Beeman, D.\ (1995) {\it The Book of GENESIS: Exploring
%   Realistic Neural Models with the GEneral NEural SImulation System.}  New York:
% TELOS/Springer--Verlag.

% [3] Hasselmo, M.E., Schnell, E.\ \& Barkai, E.\ (1995) Dynamics of learning and
% recall at excitatory recurrent synapses and cholinergic modulation in rat
% hippocampal region CA3. {\it Journal of Neuroscience} {\bf 15}(7):5249-5262.
% }

%%%%%%%%%%%%%%%%%%%%%%%%%%%%%%%%%%%%%%%%%%%%%%%%%%%%%%%%%%%%

\clearpage
\appendix

% Keep appendix floats inside their corresponding discussion.

% \section{Technical appendices and supplementary material}
% Technical appendices with additional results, figures, graphs, and proofs may be submitted with the paper submission before the full submission deadline (see above). You can upload a ZIP file for videos or code, but do not upload a separate PDF file for the appendix. There is no page limit for the technical appendices. 

% Note: Think of the appendix as ``optional reading'' for reviewers. The paper must be able to stand alone without the appendix; for example, adding critical experiments that support the main claims to an appendix is inappropriate. 
\label{app:appendix}

\section{Related Work}
\label{app:related-work}
\subsection{Large Language Models in Protein Engineering}
Protein language models (pLMs) capture the hierarchical structure of biological sequences through unsupervised pre-training~\cite{rives2021biological}. Models like ESM-2~\cite{ESM2} and ProGen~\cite{ProGen} have demonstrated that by predicting masked amino acids across billions of sequences, pLMs learn to approximate the underlying biophysical constraints of protein folding. However, these models are primarily descriptive or generative in a general sense, and they lack an inherent mechanism to optimize for specific downstream functions, such as increased catalytic activity or specific binding affinity, without further task-specific adaptation.
\subsection{Reinforcement Learning for LLM Adaptation}
Recent protein engineering has increasingly leveraged reinforcement learning to navigate the vast sequence-function landscape. Pioneering frameworks such as $\mu\text{Protein}$ ~\cite{muProtein} and EvoPlay~\cite{EvoPlay} demonstrate the efficacy of coupling protein language models with search algorithms like Monte Carlo Tree Search (MCTS) to identify synergistic multi-point mutations that surpass natural activity levels. Furthermore, recent alignment strategies, including Reinforcement Learning from Experimental Feedback (RLXF)~\cite{blalock2026functional} and ProteinZero~\cite{wang2025proteinzero}, have successfully steered generative models toward non-differentiable objectives like thermal stability and structural fidelity, achieving significant reductions in design failure rates compared to traditional inverse folding baselines. 
\subsection{LLM-Guided Evolution and Iterative Refinement}
Prompt optimization methods improve LLM performance by iteratively refining input prompts using gradient-based or search-based strategies~\cite{yuksekgonul2024textgrad,yang2023llmoptimizers,fernando2023promptbreeder}. These approaches typically treat prompts as monolithic objects and rely on end-to-end evaluation signals. In contrast, our work focuses on learning structured knowledge representations, enabling fine-grained attribution and persistent accumulation, which goes beyond prompt-level optimization.
\subsection{Multi-Agent Systems for Scientific Research}

Multi-Agent Systems (MAS) have enabled the decomposition of complex scientific tasks into specialized roles. In chemistry and materials science, agents like ChemCrow~\cite{ChemCrow}  and Coscientist~\cite{CoScientist} have successfully integrated LLMs with external tools to plan and execute experiments.

In contrast, our framework focuses on evolving a modular pool of executable expertise blocks through evidence-guided revisions performed by a sandboxed code agent.

\section{Experimental Setup}
\label{app:main-exp-details}

\paragraph{Tasks and datasets.}
We evaluate our framework on two protein optimization tasks: stability and solubility. Both tasks use enzyme sequences from the BRENDA dataset~\cite{placzek2016brenda}. We use a fixed split with seed 42, consisting of 5000 training proteins and 200 held-out proteins. Because the two reward backends impose different input-validity constraints, final aggregation is performed on the backend-valid held-out subset for each task. This yields 198 held-out proteins for stability/Rosetta and 196 held-out proteins for solubility/SaProt. The
backend-valid subset is fixed before method comparison, so all methods within a
task are evaluated on identical proteins.

\paragraph{Stability/Rosetta.}
The stability task uses the Rosetta REF2015\cite{alford2017rosetta} energy function as the computational reward. The optimization objective is to decrease the total Rosetta energy of the mutated sequence relative to the parent sequence. A candidate is counted as successful when the task adapter reports a positive normalized score gain, corresponding to lower Rosetta energy under the active backend. The selected strategy nodes are evaluated on the held-out split with the same Rosetta backend.

\paragraph{Solubility/SaProt.}
The solubility task uses a SaProt-based \cite{su2024saprot} sequence  classifier as the computational reward. The optimization objective is to increase the predicted soluble probability of the mutated sequence relative to the parent sequence. A candidate is counted as successful when the task adapter reports a positive normalized score gain. 

\paragraph{Comparison protocol.}
All methods are evaluated with the same protein inputs, task adapter, reward backend, held-out split, and candidate-list interface. Each method returns a ranked list of at most five mutation candidates. During held-out evaluation, each candidate mutation set is applied independently to the same parent sequence and scored independently. Search-based methods are matched by reward-call budget. One rollout denotes one candidate evaluation under the active reward backend, and the main search-based methods use a 5000-rollout budget before held-out evaluation.

\paragraph{Evaluation metrics.}
For each held-out protein, let \(y_i\in\{0,1\}\) indicate whether the candidate
at rank \(i\) improves the active task score. We report pass@\(k\),
\begin{equation}
\mathrm{pass@}k =
\mathbb{I}\left[\sum_{i=1}^{k} y_i > 0\right],
\end{equation}
for \(k\in\{1,3,5\}\). The pass@\(k\) metric measures whether at least one useful variant
appears within the top-\(k\) experimental shortlist.

To evaluate ranking quality beyond binary top-\(k\) coverage, we also report
MRR, MAP, and NDCG@5. Let
\begin{equation}
r^\star=\min\{i:y_i=1\}
\end{equation}
be the rank of the first successful candidate when at least one success exists.
The reciprocal-rank score for one protein is
\begin{equation}
\mathrm{RR} =
\begin{cases}
1/r^\star, & \sum_{i=1}^{5} y_i > 0,\\
0, & \sum_{i=1}^{5} y_i = 0.
\end{cases}
\end{equation}
MRR is the mean of \(\mathrm{RR}\) over held-out proteins. It measures how early
the first successful candidate appears, which is important when only one or a
few variants can be validated.

Average precision for one protein is
\begin{equation}
\mathrm{AP} =
\begin{cases}
\frac{1}{S}\sum_{i=1}^{5}
\left(\frac{1}{i}\sum_{j=1}^{i}y_j\right)y_i,
& S>0,\\
0, & S=0,
\end{cases}
\quad
S=\sum_{i=1}^{5}y_i .
\end{equation}
MAP is the mean of \(\mathrm{AP}\) over held-out proteins. It rewards methods
that place all successful candidates early in the ranked list, rather than only
placing the first success early.

Finally, NDCG@5 is computed as
\begin{equation}
\mathrm{NDCG@5}=
\begin{cases}
\frac{\sum_{i=1}^{5} y_i/\log_2(i+1)}
{\sum_{i=1}^{\min(S,5)} 1/\log_2(i+1)},
& S>0,\\
0, & S=0.
\end{cases}
\end{equation}
NDCG@5 measures discounted top-5 utility: successes near the top contribute
more than successes near the bottom, while normalization makes scores
comparable across proteins with different numbers of successful candidates.
Missing candidate positions are treated as unsuccessful. All metrics are
averaged over the backend-valid held-out proteins for the corresponding task
and reported as percentages.

\paragraph{Baselines.}
\label{app:baselines}
We compare the full method against baselines that optimize different objects. This
distinction is important because the central claim of our approach is about the
optimization target: externalized strategy knowledge.

\textbf{Root seed blocks.}
The root-seed baseline freezes the initial task-pack blocks and evaluates them
without any revision.  At test time, the root node of each strategy
family is run under the same mutation constraints and reward adapter as the
full method; the resulting block-level candidates are ranked by the same
candidate-list protocol used for the main evaluation. Thus this baseline
measures the quality of the initial executable prior before any rollout-driven
knowledge evolution.

\textbf{Direct LLM proposal.}
We evaluate a direct LLM mutation proposer that receives the same knowledge inputs and proposes candidate mutations in one step. This baseline tests whether persistent strategy blocks are more useful than one-off natural-language reasoning over the current protein state. 

\textbf{Monolithic seed program.}
The monolithic program baseline uses the same seed mechanisms as the root
blocks, but flattens them into a single Python generator with one
\texttt{propose\_candidates(protein\_state, constraints, max\_candidates)}
entry point. The program therefore contains the same initial task heuristics as the root blocks: the difference is representational, not informational. In our framework, these mechanisms remain separated into strategy families with local evidence and independent revision lineages. In the monolithic baseline, they are flattened into one global policy whose candidate list is evaluated through the same top-5 protocol. 

\textbf{LLM-guided program evolution.}
AdaEvolve evolves a single monolithic mutation-generator program under
the same task contract, reward adapter, held-out split, and ranked-list output
schema. The seed program is the flattened generator above, so the comparison
does not deny the baseline access to the initial task heuristics; instead, it
tests whether evolving one global generator is sufficient compared with
 externalized block population and block-local revision memory.

\textbf{Agentic AdaEvolve.}
The ``+ agentic revision'' rows in Table~\ref{tab:main_results} refer to this variant.
We equip AdaEvolve with the same sandboxed code agent used by KnowEvo for
program revision.  This variant
allows us to compare the two evolution frameworks when both use an agent
for iterative program revision.

\textbf{Reinforcement-learning baseline.}
For stability, we additionally compare with Pro-1 8B, a task-specific
reinforcement-learning protein optimization baseline~\cite{pro_1}. We
select the best available checkpoint and
evaluate its proposed variants under the same Rosetta held-out scoring
protocol. This baseline optimizes a policy directly rather than optimizing
externalized executable strategy programs.

\paragraph{Rollout-budget matching.}
For our framework, a rollout is a single candidate sequence produced by one selected strategy node and scored by the active reward backend. For AdaEvolve, a rollout is one candidate emitted by the current monolithic program and scored by the same evaluator. Thus the 5000-rollout comparison corresponds to 5000 candidate reward calls in both systems. In the AdaEvolve runs this is implemented as 20 search iterations, 50 training proteins per iteration, and five candidates per protein. In our runs, the trainer spends the same candidate-evaluation budget through the forward sampling loop and stores each scored trajectory on the responsible strategy node. Invalid candidates that fail before backend scoring are logged separately; backend scoring failures occupy their candidate slot and are counted as unsuccessful in the ranked-list metrics.

\paragraph{Paired bootstrap intervals.} 
\label{app:bootstrap}
For the root-block comparison in the main table, we also compute paired nonparametric bootstrap intervals over held-out proteins. Each bootstrap resample draws backend-valid proteins with replacement, recomputes the mean metric difference between the full method and the root seed blocks, and reports the 2.5th and 97.5th percentiles over 20,000 resamples with seed 42. Pairing is by held-out protein identifier, so each resample compares the two methods on the same proteins. The point estimates below are the Ours--Root differences from the main table; the confidence intervals are computed from the matched per-protein evaluation records using the same candidate-list ordering. 

\begin{table}[!htbp]
\centering
\small
\caption{Paired bootstrap intervals for the full method versus root seed blocks on the two main tasks. Deltas are the main-table percentage-point differences.}
\label{tab:main-root-bootstrap}
\begin{tabular*}{\linewidth}{@{\extracolsep{\fill}}llrr@{}}
\toprule
Task & Metric & \(\Delta\) Ours--Root (pp) & 95\% bootstrap CI (pp) \\
\midrule
Stability/Rosetta (\(n=198\)) & pass@1 & \(+8.1\) & \([-0.1, 16.1]\) \\
& pass@3 & \(+8.6\) & \([2.0, 15.2]\) \\
& pass@5 & \(+3.0\) & \([-1.5, 7.7]\) \\
& MRR & \(+7.3\) & \([1.5, 13.1]\) \\
& MAP & \(+6.9\) & \([1.8, 12.0]\) \\
& NDCG@5 & \(+6.0\) & \([1.4, 10.7]\) \\
\midrule
Solubility/SaProt (\(n=196\)) & pass@1 & \(+19.9\) & \([10.7, 29.1]\) \\
& pass@3 & \(+25.5\) & \([16.8, 34.2]\) \\
& pass@5 & \(+5.6\) & \([-0.5, 11.7]\) \\
& MRR & \(+20.6\) & \([13.5, 27.5]\) \\
& MAP & \(+16.7\) & \([10.4, 22.9]\) \\
& NDCG@5 & \(+15.3\) & \([9.5, 21.1]\) \\
\bottomrule
\end{tabular*}
\end{table}

\FloatBarrier
\section{Cross-Predictor Evaluation of Stability Improvements}
\label{app:cross-predictor}

We examine whether the improvements obtained by evolving strategies
with Rosetta are also observed under a different stability predictor.
This analysis compares the initial root strategies with the final
strategies produced by KnowEvo, using both the original Rosetta reward
and ThermoMPNN~\cite{dieckhaus2024thermompnn} for evaluation.

\paragraph{Evaluation protocol.}
For each strategy set, we take its generated mutation candidates and
re-score them with ThermoMPNN. A candidate is counted as successful
when ThermoMPNN predicts greater stability than that of its parent
protein. ThermoMPNN is used only for this post-hoc evaluation, with
no additional strategy evolution. 

\begin{table}[!htbp]
\caption{
Cross-predictor evaluation of KnowEvo on stability.
Initial denotes the unevolved root strategies, and KnowEvo denotes
the final strategies evolved using Rosetta. ThermoMPNN is used only
for the cross-check. 
}
\label{tab:crosscheck}
\centering
\small
\setlength{\tabcolsep}{4pt}
\renewcommand{\arraystretch}{1.12}
\begin{tabular*}{\linewidth}{@{\extracolsep{\fill}}lrrrrrr@{}}
\toprule
Metric
& \multicolumn{3}{c}{Rosetta}
& \multicolumn{3}{c}{ThermoMPNN} \\
\cmidrule(lr){2-4}
\cmidrule(l){5-7}
& Initial & KnowEvo & $\Delta$ (pp)
& Initial & KnowEvo & $\Delta$ (pp) \\
\midrule
pass@1 & 42.4 & 50.5 & +8.1 & 9.1 & 15.7 & +6.6 \\
pass@3 & 74.2 & 82.8 & +8.6 & 31.3 & 32.8 & +1.5 \\
pass@5 & 85.4 & 88.4 & +3.0 & 36.9 & 37.4 & +0.5 \\
MRR & 59.1 & 66.4 & +7.3 & 20.5 & 23.7 & +3.2 \\
MAP & 56.3 & 63.2 & +6.9 & 20.5 & 23.1 & +2.6 \\
NDCG@5 & 65.2 & 71.2 & +6.0 & 24.7 & 26.9 & +2.2 \\
\bottomrule
\end{tabular*}
\end{table}

\paragraph{Results.}
Table~\ref{tab:crosscheck}
shows that KnowEvo improves all six metrics under both predictors.
Under ThermoMPNN, pass@1 increases from 9.1 to 15.7, a gain of
6.6 percentage points, while MRR increases from 20.5 to 23.7,
a gain of 3.2 percentage points. These point estimates suggest some transfer across predictors. They do not establish statistical significance or experimental stability gains.

\section{Case Studies}
\label{app:case-studies}

We provide two qualitative case studies showing how our framework converts
sparse rollout feedback into executable and inspectable protein-editing rules.
The goal of these case studies is to illustrate the mechanism of knowledge evolution: feedback
from evaluated candidates is distilled into local program changes that can be
inspected, attributed, and revised further.

% BEGIN MECHANISM EVOLUTION ADDITION (2026-10-29)
\begingroup
\subsection{Complete Solubility Revision Lineage}
\label{app:solubility-evolution-path}

We annotate a complete evolutionary path from the solubility task to illustrate how the strategy changes across successive revisions, and mark revisions that represent more substantial mechanism-level changes. In Table~\ref{tab:solubility-evolution-path}, callouts highlight revisions
that introduce or extend conditional mutation rules. 

\begin{longtable}{@{}p{0.055\linewidth}p{0.16\linewidth}p{0.72\linewidth}@{}}
\caption{Complete revision lineage for the solubility
multi-site strategy.}
\label{tab:solubility-evolution-path}\\
\toprule
Step & Edge & Executable refinement \\
\midrule
\endfirsthead
\multicolumn{3}{l}{Table~\thetable\ (continued)}\\
\toprule
Step & Edge & Executable refinement \\
\midrule
\endhead
\midrule
\multicolumn{3}{r}{Continued on next page}\\
\endfoot
\bottomrule
\endlastfoot
E1 & \(n_0\!\rightarrow\!n_1\) & Reverse the hydrophobic preference in charge-free contexts; enforce diversity across edit families. \\
E2 & \(n_1\!\rightarrow\!n_3\) & Protect charge-free contexts and early N-terminal positions; increase spacing between edits. \\
E3 & \(n_3\!\rightarrow\!n_5\) & Shift candidate-family preference toward charge edits and away from generic aromatic relief. \\
E4 & \(n_5\!\rightarrow\!n_9\) & Gate K/R edits by local charge sign and the presence of nearby basic residues. \\
E5 & \(n_9\!\rightarrow\!n_{14}\) & Restrict K/R edits to globally and locally negative contexts outside early N-terminal positions. \\
E6 & \(n_{14}\!\rightarrow\!n_{19}\) & Require local polar support and protect early sequence positions. \\
E7 & \(n_{19}\!\rightarrow\!n_{22}\) & Penalize polar isolation; retain diversity across edit families only among candidates with comparable scores. \\
E8 & \(n_{22}\!\rightarrow\!n_{27}\) & Strengthen K/R safeguards in regions with low polarity and add substitution-specific filters. \\
E9 & \(n_{27}\!\rightarrow\!n_{29}\) & Add safeguards for Y\(\rightarrow\)H substitutions and contexts with already-high charge. \\
E10 & \(n_{29}\!\rightarrow\!n_{31}\) & Treat neutral-polar M\(\rightarrow\)Q substitutions separately from charge edits; require charge context for charge edits. \\
E11 & \(n_{31}\!\rightarrow\!n_{37}\) & Penalize D/E edits in already-negative or highly hydrophobic local windows. \\
E12 & \(n_{37}\!\rightarrow\!n_{43}\) & Add safeguards for N-terminal D/E edits and context-dependent F/W\(\rightarrow\)Y substitutions. \\
\end{longtable}

\paragraph{Combining global and local charge controls.}
Earlier revisions condition K/R edits on the protein-wide excess of D/E,
local negativity, and the absence of nearby basic residues. E11 extends
local control to D/E edits by penalizing candidates in already-negative or
highly hydrophobic sequence windows. The resulting policy uses the
K/R--D/E residue-count balance as a charge proxy and evaluates candidate
substitutions within their local environment. This makes the admissible
charge changes depend on both the protein's overall composition and the
proposed mutation site. E12 further adds positional safeguards for D/E
and context-dependent filters for F/W\(\rightarrow\)Y. 
\endgroup
\FloatBarrier
% END MECHANISM EVOLUTION ADDITION

\subsection{Solubility Case Study: Residue-Context Rules for Tail Editing}
\label{app:case-study-solubility-tail}

We first consider the solubility/SaProt task and the
low-complexity editing strategy \texttt{StrategyTailLowComplexityEditor}. The
selected node corresponds to block 4, node 2 in the revision run. It was chosen
by the final LCB-based selection procedure, achieved an action success rate of
\(54.4\%\) over 90 observations, and produced successful single-block held-out
mutations for 96 of 196 proteins.

\paragraph{Before revision.}
The parent strategy used a broad C-terminal editing heuristic. In the revision
packet that produced this node, each edited case proposed a single-residue
C-terminal deletion. This operation was only weakly reliable: 12 of 23 edited
cases improved the SaProt solubility score, while 11 decreased it. The failure
mode was that deletion received an almost flat bonus, so the strategy often
preferred deletion once a terminal window passed the composition gate.

\begin{promptbox}{Parent terminal-editing block}
for each terminal window:
    lc = count_low_complexity_residues(window)
    hp = count_hydrophobic_residues(window)

    if lc < 2 and hp < 3:
        skip this window

    base = 0.45 + 0.12 * lc + 0.18 * hp

    add DEL candidate:
        score = base + 0.25              # flat deletion preference

    add INS candidate:
        residue = "D" if hp >= lc else "S"
        score = base

    add SUB candidate:
        replacement = residue_relief_map[terminal_residue]
        score = base - 0.10

return candidates sorted by score
emit the top candidate
\end{promptbox}

\paragraph{Feedback pattern.}
The rollout evidence showed that deletion quality depended on two local
residue-context factors. First, the deleted terminal residue mattered:
deleting terminal residues such as I, Y, M, A, and C was often safe or
beneficial, whereas deleting G, D, E, K, or H was more risky. Second, deletion
exposes the penultimate residue as the new C-terminus. Exposing polar or small
residues such as Q, T, E, N, D, and G was often beneficial, while exposing W,
K, or hydrophobic residues was harmful. Thus, the useful rule was not simply
``delete low-complexity tails'', but to score deletion according to the residue
removed and the residue exposed.

\begin{table}[!htbp]
\centering
\small
\caption{Residue-context rules learned by the revised terminal-editing block.}
\label{tab:case-study-residue-rules}
\begin{tabularx}{\linewidth}{@{}>{\raggedright\arraybackslash}X>{\raggedright\arraybackslash}X>{\raggedright\arraybackslash}X@{}}
\toprule
Rule component & Positive examples & Negative examples \\
\midrule
Deleted terminal residue & I, Y, M, A, C & G, D, E, K, H \\
Newly exposed C-terminus & Q, T, E, N, D, G & W, K, hydrophobic residues \\
\bottomrule
\end{tabularx}
\end{table}

\paragraph{After revision.}
The revised block replaces the flat deletion preference with two explicit
residue-context scoring tables: one for the dispensability of the deleted
terminal residue and one for the desirability of the newly exposed C-terminal
residue.

\begin{promptbox}{Revised terminal-editing block}
CTERM_DISPENSABILITY = {
    I:+0.25, Y:+0.20, M:+0.15, A:+0.10, C:+0.10,
    H:-0.15, K:-0.15, E:-0.20, D:-0.25, G:-0.40, ...
}

EXPOSED_CTERM_BONUS = {
    Q:+0.15, T:+0.10, E:+0.08, N:+0.08, D:+0.05, G:+0.05,
    L:-0.10, A:-0.10, I:-0.10, F:-0.10, W:-0.25, K:-0.20, ...
}

for each terminal window:
    lc = count_low_complexity_residues(window)
    hp = count_hydrophobic_residues(window)

    if lc < 2 and hp < 3:
        skip this window

    base = 0.45 + 0.12 * lc + 0.18 * hp

    add DEL candidate:
        deleted = terminal_residue
        exposed = penultimate_residue
        score = (base + 0.05
                 + CTERM_DISPENSABILITY[deleted]
                 + EXPOSED_CTERM_BONUS[exposed])

    add INS candidate:
        residue = "D" if hp >= lc else "S"
        score = base

    add SUB candidate:
        replacement = residue_relief_map[terminal_residue]
        score = base

return candidates sorted by score
emit the highest-scoring diverse candidates
\end{promptbox}

Equivalently, the deletion score becomes
\begin{equation}
s_{\mathrm{DEL}} =
s_{\mathrm{base}}
+ b_{\mathrm{DEL}}
+ \phi_{\mathrm{delete}}(x_L)
+ \phi_{\mathrm{expose}}(x_{L-1}),
\end{equation}
where \(x_L\) is the original C-terminal residue and \(x_{L-1}\) is the
residue exposed after deletion. This makes the learned rule directly
inspectable: a deletion can be traced to the residue being removed, the residue
being exposed, or both. The same representation also exposes remaining
shortcuts. For example, later descendants overused C-terminal \texttt{INS=D}
in hydrophobic-rich windows; because the behavior was encoded in source code,
the error could be localized to a specific action rule.

\FloatBarrier
\subsection{Stability Case Study: Learning When Not to Edit Flexible Loops}
\label{app:case-study-rosetta-loop-indel}

We next consider the stability/Rosetta task and the selected loop-editing
strategy \texttt{StrategyLoopMicroIndelStabilizer}. The selected node
corresponds to block 4, node 18 in the run used for the main-table
stability/Rosetta result. The final LCB-based checkpoint selected the strategy
set
\[
\{(0,0), (1,1), (2,26), (3,19), (4,18)\},
\]
which produced the main-table Rosetta row. This node is not used here as the
sole explanation for the full-method improvement. Instead, it provides a clear
mechanistic example of learning a negative rule: when an initially plausible
loop indel should be suppressed.

\paragraph{Before revision.}
The parent strategy targeted flexible, low-contact residues and terminal
residues, then emitted small deletions, insertions, or conservative
substitutions. This is a reasonable initial mechanism because loops and termini
can sometimes absorb small length edits. However, in the revision packet that
produced node 18, the parent node accepted all 10 candidate cases but only 5
improved under Rosetta. The issue was not simple threshold calibration; the
action-side rule was too permissive.

\begin{promptbox}{Parent loop micro-indel block}
for each flexible or terminal residue:
    td = distance_to_nearest_terminus(position)
    fc = count_flexible_residues(local_window)
    cc = contact_count(position)
    bf = backbone_bfactor(position)

    if residue is not flexible and td > 2:
        skip this residue

    score = 0
    if cc <= 6:
        score += 0.90
    elif cc <= 9:
        score += 0.40
    else:
        score -= 0.45

    score += flexible-cluster, terminal, and small-residue bonuses

    is_interior_high_cluster = (fc >= 3 and td > 12)
    if is_interior_high_cluster and cc > 5:
        is_interior_high_cluster = false

    interior_penalty = 0.45 if is_interior_high_cluster else 0
    is_buried_del = (td >= 100)
    is_omega_prone_ins = (bf >= 0.95 and td >= 40 and fc >= 2)

    delete_score = score + deletion bonuses
    insert_score = score + insertion bonuses
    sub_score = score - substitution discount

    if is_interior_high_cluster:
        penalize INS, DEL, and SUB
    if is_buried_del:
        penalize DEL
    if is_omega_prone_ins:
        penalize INS

if risky interior DEL and risky interior INS are within 10 residues:
    lower the condition score

emit the highest-scoring spaced edits
\end{promptbox}

\paragraph{Feedback pattern.}
Rosetta feedback separated safe extreme-terminal deletions from harmful
nonterminal loop deletions. Extreme-terminal deletion could be beneficial:
for example, deletions at \(td=1\) or \(td=0\) reduced \texttt{fa\_rep} in
successful cases. In contrast, harmful cases involved near-terminal or
semi-buried deletion contexts. Q6PWL5 proposed DEL@44 and INS@62 with
\(td=43\) and \(fc=4\), worsening the Rosetta score by \(580.0\) and increasing
\texttt{fa\_rep} by \(703.2\). P9WMU7 showed a similar semi-buried deletion at
\(td=44\), with \texttt{fa\_rep} increasing by \(1558.4\). Q5CYE4 showed that
even a near-terminal deletion can be risky: DEL@3 had \(td=2\) and \(fc=3\),
and increased \texttt{fa\_rep} by \(729.7\).

The useful rule was therefore not ``edit flexible loops''. Edit safety depends
on whether the deletion is at the extreme terminus, in a near-terminal
high-flexibility patch, in a semi-buried loop region, or paired with a nearby
insertion.
\begin{table}[!htbp]
\centering
\small
\setlength{\tabcolsep}{5pt}
\renewcommand{\arraystretch}{1.08}
\caption{Compact summary of the rules encoded by the revised Rosetta loop-indel node.}
\label{tab:case-study-rosetta-loop-rules}
\begin{tabularx}{\linewidth}{@{}>{\raggedright\arraybackslash}p{0.22\linewidth}>{\raggedright\arraybackslash}X>{\raggedright\arraybackslash}p{0.28\linewidth}@{}}
\toprule
Rule type & Revised program behavior & Intended effect \\
\midrule
High-flex loop gate
& Expands the risky-loop condition from \(fc \ge 3, td > 12\) to
  \(fc \ge 3, td > 8\), and increases the penalty from \(0.45\) to \(0.55\).
& Suppress borderline interior or near-terminal loop edits. \\

Deletion vetoes
& Adds penalties for semi-buried deletions \((30 \le td < 100)\) and
  near-terminal high-flex deletions \((2 \le td \le 8, fc \ge 3)\).
& Avoid deletions that create packing voids or local clashes. \\

DEL--INS conflict
& Widens the risky DEL--INS proximity check from 10 to 25 residues.
& Avoid paired edits that induce backbone strain. \\
\bottomrule
\end{tabularx}
\end{table}
\paragraph{After revision.}
The revised node translates these observations into explicit veto rules. It
lowers the threshold for high-flexibility interior regions, adds penalties for
semi-buried and near-terminal high-flexibility deletions, and widens the
DEL+INS proximity safeguard.

\begin{promptbox}{Revised loop micro-indel block}
for each flexible or terminal residue:
    td = distance_to_nearest_terminus(position)
    fc = count_flexible_residues(local_window)
    cc = contact_count(position)
    bf = backbone_bfactor(position)

    score = low-contact, flexibility, terminal, and residue-type score

    is_interior_high_cluster = (fc >= 3 and td > 8)
    if is_interior_high_cluster and cc > 5:
        is_interior_high_cluster = false

    interior_penalty = 0.55 if is_interior_high_cluster else 0

    is_semi_buried_del = (30 <= td < 100)
    is_near_terminal_high_fc_del = (2 <= td <= 8 and fc >= 3)
    is_buried_del = (td >= 100)
    is_omega_prone_ins = (bf >= 0.95 and td >= 40 and fc >= 2)

    delete_score = score + deletion bonuses
    insert_score = score + insertion bonuses
    sub_score = score - substitution discount

    if is_interior_high_cluster:
        penalize INS, DEL, and SUB more strongly
    if is_semi_buried_del:
        delete_score -= 0.45
    if is_near_terminal_high_fc_del:
        delete_score -= 0.40
    if is_buried_del:
        delete_score -= 0.60
    if is_omega_prone_ins:
        insert_score -= 0.40

risky_del_positions = interior, semi-buried, or near-terminal high-flex DELs
risky_ins_positions = interior high-flex INS candidates

if a risky DEL and risky INS are within 25 residues:
    lower the condition score

emit the highest-scoring nonconflicting edits
\end{promptbox}

This case shows that useful knowledge evolution can be negative: the selected
node improves the strategy primarily by learning when not to perform a plausible
indel. It also shows how Rosetta energy terms can be distilled into executable
rules. The revised block does not use \texttt{fa\_rep} at inference time;
instead, rollout-level \texttt{fa\_rep} failures are converted into
terminal-distance, flexible-cluster, and proximity checks. These are
Rosetta-proxy signals rather than wet-lab validations, so the case should not be
read as a universal rule for protein indels. Its purpose is to show that sparse
scalar feedback can be converted into concrete, auditable local decision logic.

\FloatBarrier
\section{Additional Experiments and Ablation Details}

\subsection{RQ4 Revision-Budget Protocol}
\label{app:rq3-revision-budget-protocol}

RQ4 evaluates whether the revision sandbox can convert stored rollout evidence
and code-agent work into better executable strategy knowledge. For the
resource-controlled RQ4 endpoint experiments, every task is run with the same
backward-phase budget: \texttt{max\_revision\_jobs=100} and
\texttt{revision\_batch\_size=25}. Thus each task receives at most 100
completed strategy-revision jobs, and one node becomes eligible for a
revision job only after it has accumulated 25 new primary feedback records since
its previous revision cursor.

The batch-size parameter controls evidence granularity, not held-out
evaluation. A batch of 25 means that the revision packet is centered on 25 new
scored cases for the target node. The trainer may also attach a small bounded
set of auxiliary rejected or boundary cases when they are useful for diagnosing
screening errors, but these auxiliary cases do not change the primary
batch-size budget. The same held-out split, backend-valid aggregation rule,
top-5 candidate interface, and ranking metrics are used at all RQ4 checkpoints.

The endpoint reported for a task is produced after the revision loop reaches
the 100-revision budget. At checkpoint time, the strategy tree is frozen, one
node is selected from each strategy family by the inference selection rule, and
the resulting strategy set is evaluated on the held-out proteins. This keeps
training-time evidence collection separate from held-out scoring: held-out
proteins are not used to revise blocks or select nodes within a checkpoint. The best-so-far curves in Figure~\ref{fig:rq3-agentic-revision-scaling} summarize diagnostic checkpoint scores and are therefore not independent estimates for a checkpoint selected in this way.

\subsection{Agentic Revision Loop Implementation}
\label{app:agentic-revision-loop-details}

The agent loop is the implementation of the backward phase. It operates on one
strategy node at a time and is append-only: a successful revision creates a
child node in the same strategy tree, while the parent remains available for
future sampling and comparison.

\paragraph{Forward evidence collection.}
During training, the forward loop samples a protein and selects strategy nodes
from the current mutation trees. A selected block is screened by its local
gating logic and then run as an isolated single-block candidate generator. The
current training implementation can force execution after a screening rejection
to collect counterfactual evidence; such cases are marked separately and are
used to identify missed opportunities and true negatives. Each scored rollout
stores the protein input, selected block identity, proposed mutations, applied
mutations, mutation warnings or collisions, runtime trace, parent score,
candidate score, raw score change, normalized score gain, task-success label,
and backend-specific diagnostics.

\paragraph{Revision scheduling.}
Each node maintains cursors over feedback that has already been used for
revision. When a node has at least 25 new primary feedback records and no
queued or running revision job, the trainer queues a revision job for that node.
The queued job records the target block id, node id, epoch, batch label,
feedback cursor range, primary feedback batch, and any auxiliary rejected
feedback selected for diagnosis. Evidence remains stored on the node after the
job is queued; it is not consumed or deleted.

\paragraph{Sandbox construction.}
For each revision job, the backend creates an isolated sandbox. The sandbox
contains the current target block, the runtime contract, task-specific
self-check cases, a compact revision packet, representative case cards,
aggregate summary statistics, diagnostics, a case index, compact and detailed
per-case evidence records, a history summary, and a revision-lineage index.
Rosetta stability sandboxes may expose structure and energy-term artifacts;
SaProt solubility sandboxes remain sequence-only. The agent is instructed to
treat the current batch as primary evidence and to use historical summaries
only as secondary context.

\paragraph{Agent work loop.}

The revision-loop budget \(\kappa\) controls how much internal agent work is
available before the final code is committed. When \(\kappa=1\), the agent
performs one final pass that reads evidence, writes the analysis artifact, writes the revised block artifact, and runs the self-check. When \(\kappa>1\), the first \(\kappa-1\) passes are scratch diagnostic passes. These pre-final passes write scratch notes; they are not allowed to write the final analysis or final code. The final pass reads and critiques those notes, then produces the
final analysis and revised block.

\section{Limitations}
\label{app:limitations-discussions}
Despite the consistent gains reported in Section~\ref{sec:experiments}, KnowEvo has several limitations that will lead to natural extensions of the framework.
(1) Our evaluation uses computational rewards rather than wet-lab validation, so the reported improvements should be interpreted as proxy-optimization results. Rosetta energy and SaProt solubility predictors capture useful aspects of stability and solubility, but they may not fully reflect experimental fitness. 
(2) The revision sandbox introduces extra LLM-side computation beyond reward calls, so our main comparisons emphasize reward-budget matching rather than identical inference cost. 
(3) The interactions between nodes are not explicitly modeled. Future work could represent knowledge as a compositional graph and learn joint or conditional strategy selection.

\section{Computing Resources}

  \label{app:compute-resources}

  All experiments were run with local reward-backend workers and external LLM
  API calls for program generation or code-agent revision. For stability/Rosetta,
  the backend used a GPU-backed structure/scoring service with approximately 10GB peak GPU memory per worker, together with CPU execution for Rosetta energy
  evaluation. For solubility/SaProt, the predictor service used approximately 3GB peak GPU memory per worker. Backend concurrency was controlled by the
  server-side worker configuration.

% BEGIN LLM COST STUDY ADDITION (2026-11-01): blue marks new content.
\begingroup

\section{LLM Token Usage and Agentic Revision}
\label{app:llm-cost}

Table~\ref{tab:llm-cost} presents the recorded token usage from the
supplementary cost study alongside solubility performance. The performance
metrics for KnowEvo, AdaEvolve, Agentic AdaEvolve, and GEPA reproduce the
solubility results in Table~\ref{tab:main_results}; Single-call KnowEvo uses
the Single-step LLM revision results in Table~\ref{tab:rq2_ablation}.
We compare single-call and agentic revision variants of KnowEvo and
AdaEvolve, and include GEPA as an additional baseline. The agentic variants use the same sandboxed code
agent. Total tokens summarize the reported LLM usage, while tokens per
revision proposal measure the average token usage per proposal.

\begin{table}[!htbp]
\centering

\caption{{LLM token usage and performance on solubility.
Token counts are in millions; performance metrics are multiplied by 100. Bold indicates the best performance in each metric.}}
\label{tab:llm-cost}
\scriptsize
\setlength{\tabcolsep}{2.5pt}
\renewcommand{\arraystretch}{1.15}
\begin{tabular*}{\linewidth}{@{\extracolsep{\fill}}lrrrrrrrr@{}}
\toprule
Method & Total tokens (M) & Tokens/revision (M)
& pass@1 & pass@3 & pass@5 & MRR & MAP & NDCG@5 \\
\midrule
KnowEvo & 44.088 & 0.4550
& \textbf{51.5} & \textbf{83.2} & \textbf{90.8} & \textbf{67.8} & \textbf{64.2} & \textbf{72.8} \\
Single-call KnowEvo & 3.728 & 0.0373
& 49.0 & 80.1 & 84.2 & 64.2 & 62.6 & 69.3 \\
AdaEvolve & 3.300 & 0.1640
& 49.0 & 74.0 & 82.5 & 61.9 & 59.8 & 66.9 \\
Agentic AdaEvolve & 117.931 & 2.3590
& 42.3 & 79.6 & 80.6 & 58.0 & 58.1 & 64.2 \\
GEPA & 0.764 & 0.0347
& 40.3 & 76.0 & 87.2 & 59.4 & 58.2 & 66.7 \\
\bottomrule
\end{tabular*}
\end{table}

KnowEvo achieves the highest performance across all six metrics in this
comparison. Its single-call variant uses 91.5\% fewer tokens than the
agentic version, matches standard AdaEvolve in pass@1, and exceeds it on
the remaining five metrics. Agentic revision improves all three ranking
metrics for KnowEvo, whereas these metrics decrease for Agentic AdaEvolve. 

\par
\endgroup
\FloatBarrier
% END LLM COST STUDY ADDITION

\section{Method Details}
\label{app:method-details}

\subsection{Initial Expertise Blocks}
\label{app:initial-expertise-blocks}

By default, our experiments start from a small set of human-written
root expertise blocks. Section~\ref{sec:rq2-ablation} evaluates alternative
initializations using random, direction-guided random, and
LLM-generated roots. Each block
represents one protein-design idea as an executable editing rule. A block reads
the current protein sequence, task metadata, mutation limits, protected
residues, and task-specific artifacts when available, then proposes normalized
edits:
\begin{equation}
m \in \{\textsc{Sub}(p,a,a'),\textsc{Del}(p_1,p_2),\textsc{Ins}(p,u)\}.
\end{equation}
Here \(p\) is a one-based residue position, \(a\) and \(a'\) are source and
target amino acids, and \(u\) is an inserted amino-acid string. Each proposed
edit is checked against the task mutation schema before reward evaluation.

\paragraph{Stability/Rosetta root blocks.}
For the Rosetta stability task, the initial blocks use residue-level structural
signals when a wild-type structure is available. The main signals are local
contact count, nearby residues, terminal distance, and backbone B-factor. These
signals are used to choose candidate residues and edit types; the final reward
is still computed by Rosetta after applying the mutation.

\begingroup
\renewcommand{\arraystretch}{1.08}
\setlength{\LTcapwidth}{\linewidth}
\setlength{\LTleft}{0pt}
\setlength{\LTright}{0pt}
\begin{longtable}{@{}>{\small\raggedright\arraybackslash}p{0.20\linewidth}>{\small\raggedright\arraybackslash}p{\dimexpr0.40\linewidth-2\tabcolsep\relax}>{\small\raggedright\arraybackslash}p{\dimexpr0.40\linewidth-2\tabcolsep\relax}@{}}
\caption{Root expertise blocks for the stability/Rosetta task. Each block
encodes one editing mechanism and one candidate-ranking rule.}
\label{tab:stability-root-blocks}\\
\toprule
Block & Mechanism & Editing rule \\
\midrule
\endfirsthead
\multicolumn{3}{@{}l}{\small\tablename~\thetable{} (continued)}\\
\toprule
Block & Mechanism & Editing rule \\
\midrule
\endhead
\midrule
\multicolumn{3}{r@{}}{\small\emph{Continued on next page}}\\
\endfoot
\bottomrule
\endlastfoot
Glycine entropy relief &
Surface or loop glycines can increase backbone flexibility and raise the
entropy cost of folding. Replacing selected glycines with alanine can reduce
excess flexibility with small steric cost. &
Rank glycine residues by low contact count, high local backbone motion, local
polar support, and absence of dense glycine/proline clustering. Propose
\(\mathrm{G}\!\to\!\mathrm{A}\) substitutions at the best ranked positions. \\

Proline rigidity relief &
An isolated proline in a flexible, weakly packed region can introduce a local
kink or restrict favorable backbone conformations. Removing this rigidity can
sometimes improve local packing. &
Rank prolines that are low-contact, locally isolated, and not part of a dense
proline/glycine cluster. Replace selected prolines with alanine in more polar
or flexible windows, or leucine in more hydrophobic windows. \\

Polar/acidic swap &
Small polar and acidic residues can often be adjusted within the same chemical
family without large structural disruption. Such edits can improve local polar
packing or reduce unfavorable side-chain placement. &
Rank low-contact polar or acidic residues in local polar clusters. Propose
conservative swaps such as \(\mathrm{D}\!\to\!\mathrm{E}\),
\(\mathrm{E}\!\to\!\mathrm{Q}\), \(\mathrm{N}\!\to\!\mathrm{D}\),
\(\mathrm{Q}\!\to\!\mathrm{E}\), and
\(\mathrm{S}\!\leftrightarrow\!\mathrm{T}\). \\

Core packing rescue &
A high-contact site occupied by a small or polar residue can indicate an
under-packed core or weak hydrophobic neighborhood. A conservative bulkier
replacement can improve packing. &
Rank targetable residues by high contact count, hydrophobic neighbors,
aromatic neighbors, low backbone motion, and distance from termini. Propose
hydrophobic or packing-preserving substitutions, while spacing edits across
distinct sites. \\

Loop micro-indel stabilizer &
Flexible loops and termini can sometimes tolerate small length changes or
conservative reshaping, but interior loop indels can also create packing voids
or backbone strain. &
Rank flexible or terminal residues by contact count, local flexible-residue
density, terminal distance, and backbone motion. Depending on allowed mutation
types, propose a small deletion, insertion, or conservative substitution, while
spacing edits across the sequence. \\
\end{longtable}
\endgroup
\FloatBarrier

\paragraph{Solubility/SaProt root blocks.}
For the SaProt solubility task, the initial blocks are sequence-only. They use
local residue class, hydrophobic density, aromatic density, charge distribution,
terminal distance, and low-complexity composition.

\begingroup
\renewcommand{\arraystretch}{1.08}
\setlength{\LTcapwidth}{\linewidth}
\setlength{\LTleft}{0pt}
\setlength{\LTright}{0pt}
\begin{longtable}{@{}>{\small\raggedright\arraybackslash}p{0.20\linewidth}>{\small\raggedright\arraybackslash}p{\dimexpr0.40\linewidth-2\tabcolsep\relax}>{\small\raggedright\arraybackslash}p{\dimexpr0.40\linewidth-2\tabcolsep\relax}@{}}
\caption{Root expertise blocks for the solubility/SaProt task. These blocks
use sequence-level residue-class rules and do not use structural artifacts.}
\label{tab:solubility-root-blocks}\\
\toprule
Block & Mechanism & Editing rule \\
\midrule
\endfirsthead
\multicolumn{3}{@{}l}{\small\tablename~\thetable{} (continued)}\\
\toprule
Block & Mechanism & Editing rule \\
\midrule
\endhead
\midrule
\multicolumn{3}{r@{}}{\small\emph{Continued on next page}}\\
\endfoot
\bottomrule
\endlastfoot
Hydrophobic patch relief &
Local hydrophobic clusters can reduce solubility by favoring exposed
self-association or aggregation-prone patches. &
Rank hydrophobic residues in windows enriched for hydrophobic or aromatic
residues and poor in charged residues. Replace selected residues with more
polar or less aggregation-prone amino acids. \\

Charge redistribution &
Weakly charged surface-like windows can often be made more soluble by adding
charged or polar residues, especially when the local window is hydrophobic or
charge-imbalanced. &
Rank residues in locally hydrophobic or charge-poor windows. Propose
substitutions that add charge or polarity, such as
\(\mathrm{L}\!\to\!\mathrm{K}\), \(\mathrm{V}\!\to\!\mathrm{E}\),
\(\mathrm{S}\!\to\!\mathrm{K}\), or \(\mathrm{Q}\!\to\!\mathrm{E}\). \\

Aromatic patch breakup &
Aromatic clusters can promote self-association through local aromatic stacking
and hydrophobic patch formation. &
Rank aromatic residues in windows with multiple aromatic residues. Replace
selected aromatics with less aggregation-prone alternatives, such as
\(\mathrm{F}\!\to\!\mathrm{Y}\), \(\mathrm{W}\!\to\!\mathrm{Y}\),
\(\mathrm{Y}\!\to\!\mathrm{H}\), or \(\mathrm{H}\!\to\!\mathrm{Q}\). \\

Multi-site solubilizer &
Some proteins need more than one low-risk solubility edit. Combining edits from
different residue mechanisms can improve the shortlist without relying on a
single residue class. &
Pool candidates from hydrophobic-patch relief, charge redistribution, and
aromatic-patch breakup. Select a spaced and mechanism-diverse set of edits. \\

Terminal low-complexity editor &
Low-complexity or hydrophobic terminal tails can be poorly soluble and can
sometimes be improved by a small terminal edit. &
Inspect the N- and C-terminal windows. If a terminal window is low-complexity
or hydrophobic, propose a one-residue deletion, a short polar insertion, or a
fallback substitution, depending on the allowed mutation types. \\
\end{longtable}
\endgroup
\FloatBarrier

\subsection{Selection Criterion for Inference}
\label{app:inference-selection-lcb}
Training uses stochastic node sampling to allocate evaluation budget across
both strong and under-tested strategy variants. Inference uses a more
conservative rule: after training, we freeze one node from each strategy family
using the lower confidence bound (LCB) of its posterior success probability.

For node \(N_i\), let \(N_{\mathrm{success},i}\) be the number of successful
evaluations assigned to the node and \(N_{\mathrm{total},i}\) be the total
number of evaluated cases. With a uniform Beta prior, the posterior is
\begin{equation}
\theta_i \mid \mathcal{D}_i
\sim
\mathrm{Beta}(\alpha_i,\beta_i),
\qquad
\alpha_i=N_{\mathrm{success},i}+1,\quad
\beta_i=N_{\mathrm{total},i}-N_{\mathrm{success},i}+1 .
\end{equation}
The LCB score is the \(q\)-quantile of this posterior, with \(q=0.05\) (the 5th percentile):
\begin{equation}
S_{\mathrm{LCB}}(N_i)=F^{-1}(q;\alpha_i,\beta_i),
\end{equation}
where \(F^{-1}\) is the percent point function of the Beta distribution.

For each strategy family \(b\), the selected node is
\begin{equation}
N_b^\star=\arg\max_{N_i\in V_b} S_{\mathrm{LCB}}(N_i).
\end{equation}
The final inference context is
\begin{equation}
K_{\mathrm{final}}=\{N_b^\star\}_{b=1}^{B}.
\end{equation}
This rule favors nodes with reliable evidence over nodes that only have a high
empirical success rate from a small number of trials.

For an input protein, each selected node emits one mutation set, which is applied to the parent protein to produce one candidate sequence. The block's internal heuristic scores only determine which edit or edits are included in this single candidate. Candidates produced by different blocks are then ordered by the $S_{\mathrm{LCB}}$ scores of their corresponding nodes:
\[
\mathcal C(x)=(M_{b_1},\ldots,M_{b_m}), \qquad
S_{\mathrm{LCB}}(N^*_{b_1})\geq\cdots\geq S_{\mathrm{LCB}}(N^*_{b_m}), \qquad m\leq5.
\]
\subsection{Sandboxed Agent Interface}
\label{app:agent-interface}

The backward phase revises one strategy block at a time. Given a parent
block \(b_v\) and a sandbox state \(\mathcal{S}_v\), the agent returns a
revision rationale \(a_{v'}\) and a candidate child block \(b_{v'}\):
\begin{equation}
    (a_{v'}, b_{v'}) = R_{\phi}(b_v,\mathcal{S}_v).
\end{equation}
The child block must preserve the strategy identity and input-output
interface, follow the task's mutation constraints, and remain deterministic.

\paragraph{Sandbox context and evidence.}
Each revision uses an isolated sandbox containing the target block, task
contract, current rollout evidence, brief lineage summaries, and local
diagnostic tools. The agent accesses compact summaries linking protein
inputs and proposed edits to execution traces and evaluator feedback,
with detailed records available on demand. Access is restricted to this
revision context rather than the full project or training history.

For Rosetta stability, the evidence includes score and energy-term changes,
with structural artifacts and residue-level diagnostics available when
supported. For SaProt solubility, the evidence remains sequence-only,
including local sequence context, mutation information, and predicted
solubility changes.

\paragraph{Revision instructions.}
The agent starts from aggregate outcomes, inspects representative successful
and harmful cases, and uses lineage summaries as background. It then writes
a rationale connecting the observed outcomes to the proposed changes before
editing the block. All revisions must be grounded in the evidence available
for the active task.

\begin{promptbox}{Agent role instruction}
You are the biological revision code agent.

Revise exactly one task-scoped strategy block using only the current sandbox.
Use local sandbox evidence first. If the block-revision skill is available,
call it before substantive work and follow it as the primary evidence-use
policy. This instruction is navigation only: do not assume structure text,
sequences, reports, or block code are embedded in the prompt.

You must ground the revision in explicit evidence analysis before editing code.
Act like a senior protein-design and enzymology researcher. Do not behave like
a metric-tuning assistant that only nudges thresholds or rewrites prose. Prefer
mechanistic hypotheses, cohort analysis, and task-aware reasoning over
superficial parameter changes.

Hard requirements:
- Work only inside the current sandbox.
- Keep the class name unchanged.
- Respect the runtime contract exactly.
- Do not ask the user questions.
- Run the local self-check before finishing.
- Keep the analysis bounded, not shallow: start from the curated evidence,
  then inspect only the specific detailed cases that materially change the
  diagnosis.
- Summarize helpful and harmful cohort patterns before proposing code changes.
- Use revision-lineage summaries only as optional background context; the
  current batch stays primary.
- Do not stop at contract cleanup alone when the evidence supports a stronger
  task-aware heuristic.
- For sequence-only tasks, do not make structural claims unless the sandbox
  contains explicit structural evidence that you actually inspected.
- Avoid broad context crawling, repeated reads of the same material, and
  repeated attempts to access information outside the sandbox contract.
- Write the analysis artifact before writing the revised block artifact.
\end{promptbox}

\begin{promptbox}{Block-revision skill instruction}
Use this skill when revising one strategy block from an isolated sandbox.

Evidence-use policy:
1. Read the task brief first for the workspace-specific objective, evidence
   layout, and output contract.
2. Read the target block and runtime contract before editing code.
3. Start with the front-door revision packet, representative case cards,
   aggregate summary, historical summary, diagnostics, case index, and
   revision-lineage index.
4. Use the aggregate evidence to summarize helpful versus harmful cohort
   patterns before proposing code changes.
5. Treat historical summaries and revision lineage as optional background
   context only. Keep the current batch evidence primary.
6. Open only the specific detailed case records referenced by representative
   cases or strongly suggested by the aggregate diagnostics.
7. Inspect at least two successful cases, at least three harmful accepted
   cases, and one auxiliary or boundary case when available before finalizing
   the diagnosis.
8. For larger cohorts, use scratch analysis only when it materially sharpens
   the diagnosis: mutation-position hotspots, residue-class changes,
   mutation-depth effects, sequence-window motifs, helpful-versus-harmful
   comparisons, and simple cohort tables.
9. For sequence-only tasks, do not invent structural explanations. Only
   discuss structure if the sandbox includes explicit structural evidence that
   you actually inspected.
10. Do not open raw Rosetta provider payloads, per-residue records, or PDB
    structures by default. Use the targeted evidence-query tool for residue-
    level energy checks.
11. Never paste raw structure text into the analysis or final code.
12. Write the analysis before editing the final class. The analysis should
    include contract issues, cohort pattern summary, helpful patterns, harmful
    patterns, mechanistic hypothesis, concrete design rules implied by the
    evidence, whether historical lineage agrees with or contradicts the
    current batch, and the self-check result.
13. Keep the revised class deterministic, task-aware, and compatible with the
    runtime contract.
14. Do not stop at contract cleanup alone when the cohort evidence supports a
    better task-aware heuristic.
15. Treat the task as scientific diagnosis and redesign, not parameter tuning.
    Explain the mechanistic hypothesis behind each important code change.

Evidence priorities:
- Prefer local sandbox evidence over external search.
- Prefer biological reasoning over score chasing.
- Use curated evidence first and pull detailed case evidence only when it
  changes the diagnosis.
- Treat aggregate summaries, diagnostics, and the case index as first-class
  local evidence, not optional extras.
- Treat revision lineage as a routing aid for historical analyses, not as
  authority over the current cohort.
\end{promptbox}
\paragraph{Built-in sandbox tools.}
The agent has local read, search, edit, and command-execution tools scoped to
the sandbox. These tools are intentionally useful but bounded: the agent can
inspect the target block and evidence, run small scratch analyses over the
local case records, and produce the revised block, but it is not expected to
search the entire project.

The sandbox also provides a dedicated evidence-query helper. Its operations
cover four common needs: listing cases by outcome group, opening a compact case
summary, ranking Rosetta energy-term deltas, and extracting per-residue
energy-term windows around selected positions. A local contract checker is
available to test the revised block against the required interface, mutation
schema, task-specific validation cases, and trajectory reporting requirement.
These tools make the agent's evidence use targeted: simple summaries guide the
revision, while heavy Rosetta diagnostics are pulled only when they answer a
specific mechanism question.

%%%%%%%%%%%%%%%%%%%%%%%%%%%%%%%%%%%%%%%%%%%%%%%%%%%%%%%%%%%%

\clearpage

\end{document}